\documentclass{article}

\usepackage[final]{neurips_2022}

\pdfoutput=1
\usepackage[utf8]{inputenc} % allow utf-8 input
\usepackage[T1]{fontenc}    % use 8-bit T1 fonts
\usepackage{booktabs}       % professional-quality tables
\usepackage{amsfonts}       % blackboard math symbols
\usepackage{nicefrac}       % compact symbols for 1/2, etc.
\usepackage{microtype}      % microtypography
\usepackage{xcolor}        
\usepackage{tikz}
\usepackage{pgfplots}
\usepackage{xcolor}
\usepackage{amssymb}
\usepackage{wrapfig}
\usepackage{lipsum}

\usepackage{caption}
\usepackage{subcaption}

\usepackage{filecontents}
\usepackage{algorithm}
\usepackage{algorithmic}
\usepackage{algorithm}
\usepackage{wrapfig}
\usepackage{lipsum}
\usepackage{amsmath}
\usepackage{graphicx}
\usepackage{url}
\usepackage{amsfonts}
\usepackage{color, soul}
\usepackage{mathrsfs}
\usepackage{cleveref}
\usepackage{multirow}
\usepackage{float}
\usepackage{wrapfig}
\usepackage{amsthm}
\usepackage{bm}
\usepackage{bbm}
\usepackage{algorithm}
\usepackage{algorithmic}
\usepackage{colortbl}
\usepackage{enumitem}

\newtheorem{definition}{Definition}
\newtheorem{assumption}{Assumption}

\usepackage{multirow}
\title{CLEAR: Generative Counterfactual Explanations \\on Graphs}

\author{%
  Jing Ma \\ % \thanks
  University of Virginia\\
  Charlottesville, VA, USA \\
  \texttt{jm3mr@virginia.edu} \\
  \And
  Ruocheng Guo \\
  Bytedance AI Lab \\
  London, UK \\
  \texttt{rguo.asu@gmail.com}
  \And
  Saumitra Mishra \\
  J.P. Morgan AI Research \\ 
  London, UK \\
  \texttt{saumitra.mishra@jpmorgan.com}
  \And
  Aidong Zhang \\
  University of Virginia\\
  Charlottesville, VA, USA \\
  \texttt{aidong@virginia.edu} \\
  \And
  Jundong Li \\
  University of Virginia\\
  Charlottesville, VA, USA \\
  \texttt{jundong@virginia.edu} \\
}

\begin{document}

\maketitle

\newcommand{\mymodel}{CLEAR}
\newcommand{\bigCI}{\mathrel{\text{\scalebox{1.07}{$\perp\mkern-10mu\perp$}}}}
\begin{abstract}
%Recently, a surge of research about machine learning (ML) on graphs has sprung up and  stimulated the deployment of ML prediction models in real-world decision systems. However, such decisions often lack human-understandable explanations. 
%Among existing studies of machine learning model explainability, %explainable artificial intelligence (XAI), 
Counterfactual explanations promote explainability in machine learning models by answering the question ``\emph{how should an input instance be perturbed to obtain a desired predicted label?}". The comparison of this instance before and after perturbation can enhance human interpretation. Most existing studies on counterfactual explanations are limited in tabular data or image data. 
In this work, we study the problem of counterfactual explanation generation on graphs. %in a model-agnostic setting. 
A few studies have explored counterfactual explanations on graphs, but many challenges of this problem are still not well-addressed: 1) optimizing in the discrete and disorganized space of graphs; 2) generalizing on unseen graphs; and 3) maintaining the causality in the generated counterfactuals without prior knowledge of the causal model.
To tackle these challenges, we propose a novel framework \mymodel~which aims to generate counterfactual explanations on graphs for graph-level prediction models. Specifically, \mymodel~leverages a graph variational autoencoder based mechanism to facilitate its optimization and generalization, and promotes causality by leveraging an auxiliary variable to better identify the underlying causal model. Extensive experiments on both synthetic and real-world graphs validate the superiority of \mymodel~over the state-of-the-art methods in different aspects. %Notably, \mymodel~ outperforms state-of-the-art baselines in efficient generalization and maintaining causality. 
%All the code and dataset are available through the link in the Appendix B.

%efficiently generate (multiple) counterfactuals for each graph, and easily be generalized to unseen graphs. Furthermore, \mymodel~ significantly outperforms baselines in the causality of counterfactuals.
%Extensive experiments on both synthetic and real-world graphs validate the superiority of our framework in different aspects. 
\end{abstract}


%\keywords{sentiment analysis}

\maketitle

%\vspace{-3mm}
\section{Introduction}
\begin{wrapfigure}{l}{0.45\textwidth}
%\vspace{-.8cm}
  \begin{center}
    \includegraphics[width=0.45\textwidth]{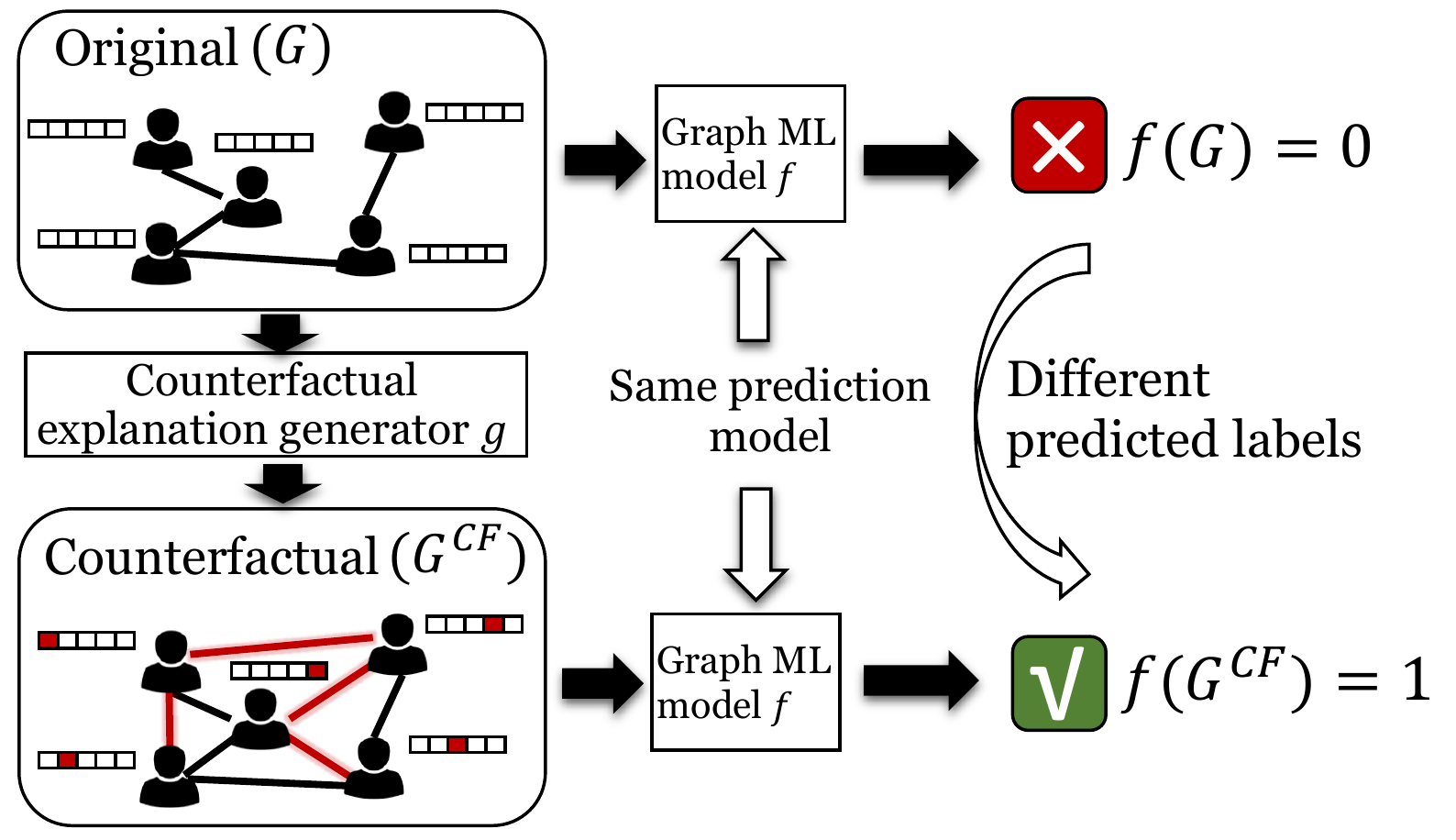} 
 %   \vspace{-.6cm}
  \end{center}
  \caption{An example of CFE on graphs. %Given a prediction model $f$, we seek an algorithm $g$ to generate counterfactuals for any input graph $G$. A counterfactual $G^{CF}$ is similar to $G$, but has a different label predicted by $f$.
  } 
%\vspace{-.5cm}
\label{fig:example1}
\end{wrapfigure}

To facilitate explainability in opaque machine learning (ML) models (e.g., how predictions are made by a model, and what to do to achieve a desired outcome), %produce explainable models or justifications for model predictions, 
explainable artificial intelligence (XAI) \cite{linardatos2020explainable} has recently attracted significant attention in many communities. Among the existing work of XAI, a special class, i.e., \textit{counterfactual explanation} (CFE) \cite{wachter2017counterfactual}, promotes model explainability by answering the following question: ``\emph{For a specific instance, how should the input features $X$ be slightly perturbed to new features $X'$ to obtain a different predicted label (often a desired label) from ML models?}". 
%More formally, for a data instance whose prediction made by the model needs to be explained (this instance is called an \textit{explainee instance}), CFE studies aim to generate new data instances which have similar features to the explainee instance, but can lead to predicted labels different from that of the explainee instance.
%
The original instance whose prediction needs to be explained is called an \textit{explainee instance}, and the generated instances after perturbation are referred to as ``counterfactual explanations".
Generally, CFE promotes human interpretation through the comparison between $X$ and $X'$. With its intuitive nature, CFEs can be deployed in various real-world scenarios such as loan application and legal framework \cite{verma2020counterfactual}. Different from traditional CFE studies \cite{wachter2017counterfactual,verma2020counterfactual,mahajan2019preserving,mishra2021survey} on tabular or image data, recently, CFE on graphs is also an emerging field in many domains with graph structure data such as molecular analysis \cite{numeroso2021meg} and professional networking \cite{wolff2009effects}. 
%Notice that CFE is different from adversarial attack. As discussed in \cite{bajaj2021robust}, adversarial attack focuses only on changing the predicted labels but completely neglects the explainability of the adversarial examples.
%Differently, CFEs can benefit model explainability in various aspects and deployment scenarios.
%
%For example, for a job application prediction model with each input instance as a graph representing an applicant's career networking, applicants can utilize the counterfactuals corresponding to a desired label (being hired) to modify their profiles as well as establishing better networking, and applicants can also report biases if the counterfactuals involve perturbations on certain sensitive attributes such as race or gender. 
For example, consider a grant application prediction \cite{grant} model with each input instance as a graph representing a research team's collaboration network, where each node represents a team member, and each edge signifies a collaboration relationship between them.
Team leaders can improve their teams for next application by changing the original graph according to the counterfactual with a desired predicted label (application being granted). If the counterfactual is more dense than the original, the team leader may then encourage more team collaborations. %Furthermore, teams can  report discrimination if the counterfactuals involve perturbations on certain sensitive attributes such as race or gender of team members. 
%In a molecular property prediction task where each instance is a molecular graph, researchers can use the counterfactuals to discover new compounds with a desired property, and can also refine the prediction model by checking whether the predictions are  consistent with domain knowledge. 
%In the above job application example, 
%Currently, most existing studies of CFEs \cite{wachter2017counterfactual,verma2020counterfactual,mahajan2019preserving,mishra2021survey} are based on either tabular or image-based data. 
To this end, in this work, we investigate the problem of generating counterfactual explanations on graphs. As shown in Fig.~\ref{fig:example1}, given a prediction model $f$ on graphs, for a graph instance $G$, we aim to generate counterfactuals (e.g., $G^{CF}$) which are slightly different from $G$ w.r.t. their node features or graph structures to elicit a desired model prediction. Specifically, we focus on graph-level prediction without any assumptions of the prediction model type and its model access, i.e., $f$ can be a black box with unknown structure.

% \begin{wrapfigure}[13]{L}{0.55\textwidth}
% \vspace{-8mm}
% \begin{minipage}{0.55\textwidth}
% \begin{figure}
% \centering
%         \includegraphics[width=.41\textwidth,height=1.7in]{figures/graphCFE_motivation_cropped.pdf}
%         \vspace{-0.12in}
%         \caption{An illustrative example of counterfactual explanations on graphs. Given a prediction model $f$ on graphs, we seek an algorithm $g$ to generate counterfactuals for any input graph $G$. A counterfactual $G^{CF}$ is a graph that is similar to $G$, but has a different label (often a desired one) predicted by $f$.} 
%         \vspace{-0.16in}
%          \label{fig:example1}
%  \end{figure}
%  \end{minipage}
% \end{wrapfigure}
 
% limitations of previous works & challenges
Recently, a few studies \cite{lucic2021cf,numeroso2021meg,bajaj2021robust,abrate2021counterfactual,ying2019gnnexplainer,yuan2020explainability} explore to extend CFEs into graphs. However, this problem still remains a daunting task due to the following key challenges: 
1) \textbf{Optimization:} Different from traditional data, the space of perturbation operations on graphs (e.g., add/remove nodes/edges) is discrete, disorganized, and vast, which brings difficulties for optimization in CFE generation. 
Most existing methods \cite{numeroso2021meg,abrate2021counterfactual} %formulate CFE generation as a reinforcement learning problem, and
search for graph counterfactuals by enumerating all the possible perturbation operations on the current graph.
%in an iterative way: at each step, they enumerate all the possible perturbation operations on the current graph to form a space and select one from them until a desired label is obtained. 
However, such enumeration on graphs is of high complexity, % to enumerate all the perturbation operations on graphs. 
and it is also challenging to solve an optimization problem in such a complex search space. 
Few graph CFE methods which enable gradient-based optimization either rely on domain knowledge \cite{numeroso2021meg} or assumptions \cite{bajaj2021robust} about the prediction model to facilitate optimization. However, these knowledge and assumptions limit their applications in different scenarios.
%seek for the direction of optimization in such a disorganized space. 
%Thus they often rely on specific domain knowledge \cite{numeroso2021meg} (e.g., chemical rules) or assumptions \cite{bajaj2021robust} about the prediction model $f$ (e.g., $f$ is a graph neural network (GNN) model and its gradients or representations are accessible) to prune the search space and facilitate the optimization.
%and some other methods 
%However, such knowledge limit their application in different domains. 
%Some other approaches of CFE on graphs 
%\cite{bajaj2021robust,numeroso2021meg} rely on the assumption about the type and access of the prediction model (e.g., the prediction model is a GNN model, or certain parts of it, such as the gradients or representations, are accessible) to facilitate optimization. 
%However, these domain knowledge and assumptions limit their application in different scenarios.
%
2) \textbf{Generalization:} The discrete and disorganized nature of graphs also brings challenges for the generalization of CFE methods on unseen graphs, %As with such nature, 
as it is hard to sequentialize the process of graph CFE generation and then generalize it. 
Most existing CFE methods on graphs  \cite{lucic2021cf,ying2019gnnexplainer} solve an optimization problem for each explainee graph separately. %i.e., the CFE generation process is an optimization task for each specific graph instance. 
These methods, however, cannot be generalized to new graphs.
%limits the generalibility of these methods for unseen graphs. 
%Several recent studies \cite{bajaj2021robust} start to utilize the training data to learn some robust logic for CFE generation, but they are often limited to a specific type of perturbation (e.g., remove edges), thus cannot be generalized to graphs which need more complex perturbations in their  counterfactuals.
%to learn an algorithm which identifies important edges for prediction, and can generate counterfactuals for unseen graphs by removing their important edges. 
%but a drawback of such methods is that they can only remove edges, thus can not be generalized to graphs which need more complex perturbations in counterfactuals, e.g., attributed graphs, or sparse graphs which need to insert nodes/edges for counterfactuals. 
%
%A recent exploration \cite{bajaj2021robust} 
3) \textbf{Causality:} It is challenging to generate %high-quality 
counterfactuals that are consistent with the underlying causality. Specifically, causal relations may exist among different node features and the graph structure. 
%For example, in the aforementioned job application example, for each applicant, establishing more career networking in their professional domain necessitates increasing  the work experience of this applicant.
In the aforementioned example, for each team, after establishing more collaborations, the team culture may be causally influenced. %(e.g., improve the member engagement)
%A realistic counterfactuals should
%In this example, typical CFE generators often only change the collaboration numbers without changing the team culture correspondingly, which violates the causality. 
Incorporating causality can generate more realistic and feasible 
%and actionable 
counterfactuals \cite{mahajan2019preserving}, but most existing CFE methods either cannot handle causality, or require too much prior knowledge about the causal relations in data.
%A few studies on non-graph data \cite{mahajan2019preserving,karimi2020algorithmic,karimi2021algorithmic} have investigated to promote causality in counterfactuals by incorporating explicit prior knowledge of the causal relations into their CFE generators. 
%However, such prior knowledge is often unavailable in the real world. Besides, the (unknown) causal relations on graphs are often more complicated and difficult to be discovered than traditional data. % due to the connection in the graphs. 

%, but these works are not based on graph data, while on graphs, many factors relevant to the graph structure are difficult to be extracted and involved in causal constraints. Besides, the ground-truth causal relations among different variables are often unknown in real world, which makes it difficult to capture and enforce causality in CFEs. Existing works \cite{mahajan2019preserving,karimi2020algorithmic,karimi2021algorithmic}, however, are limited in the prerequisite of known prior knowledge of the causal relations in data.
%a prerequisite of designing causal constraint i is accessible prior knowledge of the causal relations among different variables (e.g., a causal model which describes the causal relations). However, 

% our method
To address the aforementioned challenges, in this work, we propose a novel framework --- generative \textbf{C}ounterfactua\textbf{L} \textbf{E}xpl\textbf{A}nation gene\textbf{R}ator for graphs (\mymodel). At a high level, \mymodel~is a generative, model-agnostic CFE generation framework for graph prediction models. For any explainee graph instance, \mymodel~aims to generate counterfactuals 
with slight perturbations on the explainee graph to elicit a desired predicted label, and the counterfactuals are encouraged to be in line with the underlying causality. 
More specifically, to facilitate the optimization of the CFE generator, %without iteratively enumerating graph perturbation operations, 
we map each graph into a latent representation space, and output the counterfactuals as a probabilistic fully-connected graph with node features and graph structure similar as the explainee graph. In this way, the framework is differentiable and enables gradient-based optimization. 
To promote generalization of the CFE generator on unseen graphs, we propose a generative way to construct the counterfactuals. %Specifically, \mymodel~ follows a mechanism of graph variational autoencoder (VAE) \cite{simonovsky2018graphvae}, which includes an encoder to map input explainee graphs into a latent space, and a decoder to generate counterfactuals. 
After training the CFE generator, it can be efficiently deployed to generate (multiple) counterfactuals on unseen graphs,  
%In this way, the gradient-based optimization becomes feasible because the latent space is continuous and differentiable. 
%Another advantage of this generative mechanism is that 
%It can be generalized to out-of-distribution instances, 
rather than retraining from scratch. 
%Furthermore, compared with existing methods, our method can be generalized to graphs which need more complicated perturbations for counterfactuals (e.g., add/remove nodes or edges, and modify node features) thanks to our generative mechanism.
%
To generate more realistic counterfactuals without explicit prior knowledge of the causal relations, inspired by the recent progress in nonlinear independent component analysis (ICA) \cite{khemakhem2020variational} and its connection with causality \cite{pearl2009causality}, we make an exploration to promote causality in counterfactuals by leveraging an auxiliary variable to better identify the latent causal relations.
%improving the identifiability of the CFE generator. %promote the generated CFEs to satisfy causality buried in the data
 %we build connection between ICA and the structural causal model (SCM) \cite{pearl2009causality} which describes the causal relations among different variables, then promote the generated CFEs to satisfy causality buried in the data, without explicit prior knowledge of the causal model.
The main contributions of this work can be summarized as follows:
%\begin{itemize}
1) \textbf{Problem.} We study an important problem: 
     counterfactual explanation generation on graphs. We analyze its challenges including optimization, generalization, and causality. % in counterfactual explanation generation on graphs without explicit prior knowledge of the causal model.
    To the best of our knowledge, this is the first work jointly addressing all these challenges of this problem.
2) \textbf{Method.} We propose a novel framework \mymodel~to generate counterfactual explanations for graphs. \mymodel~can generalize to unseen graphs, and promote causality in the counterfactuals.
3) \textbf{Experiments.} We conduct extensive experiments on both synthetic and real-world graphs to validate the superiority of our method over state-of-the-art baselines of graph CFE generation.

\section{Preliminaries}
%\vspace{-2mm}
A graph $G=(X,A)$ is specified with its node feature $X$ and adjacency matrix $A$. 
We have a graph prediction model $f: \mathcal{G} \rightarrow \mathcal{Y}$, where $\mathcal{G}$ and $\mathcal{Y}$ represent the space of graphs and labels, respectively. In this work, we assume that we can access the prediction of $f$ for any input graph, but we do not assume the access of any knowledge of the prediction model itself.
For a graph $G\in \mathcal{G}$, we denote the output of the prediction model as $Y=f(G)$. 
A counterfactual $G^{CF}=(X^{CF},A^{CF})$ is expected to be similar as the original explainee graph $G$, but the predicted label for $G^{CF}$ made by $f$ (i.e., $Y^{CF}=f(G^{CF})$) should be different from $f(G)$. With a desired label $Y^{*}$ (here $Y^{*}\ne Y$), the counterfactual $G^{CF}$ is considered to be \textit{valid} if and only if $Y^{*}=Y^{CF}$. In this paper, we mainly focus on graph classification, but our framework can also be extended to other tasks such as node classification, as discussed in Appendix D.
%In this paper, we use non-bold, italicized, and capitalized letters (e.g., $G$) to denote random variables; non-bold lowercase letters (e.g., $y_i$) to denote observed values of a scalar; bold capitalized letters (e.g., $\mathbf{G}$) to denote the observed values of a set/matrix.

Suppose we have a set of graphs sampled from the space $\mathcal{G}$, and different graphs may have different numbers of nodes and edges. A counterfactual explanation generator can generate counterfactuals for any input graph $G$ w.r.t. its desired predicted label $Y^*$. % Here, each counterfactual explanation $G^{CF}$ is expected to be similar as the original graph $G$, but the predicted label for $G^{CF}$ made by $f$ should be different from the predicted label $f(G)$ for the original graph $G$. With a desired label $Y^{CF}$, the generated CFE $G^{CF}$ is considered to be \textit{valid} if and only if $Y^{CF}=f(G^{CF})$.
As aforementioned, most existing CFE methods on graphs have limitations in three aspects: optimization, generalization, and causality. 
Next, we provide more background of causality. The causal relations between different variables (e.g., node features, degree, etc.) in the data can be described with a structural causal model (SCM):
%\vspace{-1mm}
\begin{definition}
(Structural Causal Model) A structural causal model (SCM) \cite{pearl2009causality} %consists of a \textit{causal graph} and a set of \textit{structural equations}. A causal graph is a directed acyclic graph (DAG) using directed edges to denote causal relations among variables. SCM 
is denoted by a triple $(U,V,F)$: $U$ is a set of exogenous variables, and $V$ is a set of endogenous variables. The structural equations $F\!=\!\{F_1,...,F_{|V|}\}$ determine the value for each $V_i\!\in\! V$ with $V_i \!=\! F_i(\mathtt{PA}_i, U_i)$, here $\mathtt{PA}_i\!\subseteq\!{V}\backslash V_i$ denotes the ``parents" of $V_i$, and $U_i\!\subseteq\! U$.% denotes the parent exogenous variables of $V_i$. %The causal relations in a SCM can often be illustrated with a directed acyclic graph (DAG), where  variables are represented with nodes, and causal relations are represented with directed edges between nodes. This graph is known as causal graph.
\end{definition}
%\vspace{-2mm}

\begin{wrapfigure}{l}{0.44\textwidth}
%\vspace{-.8cm}
  \begin{center}
    \includegraphics[width=0.44\textwidth]{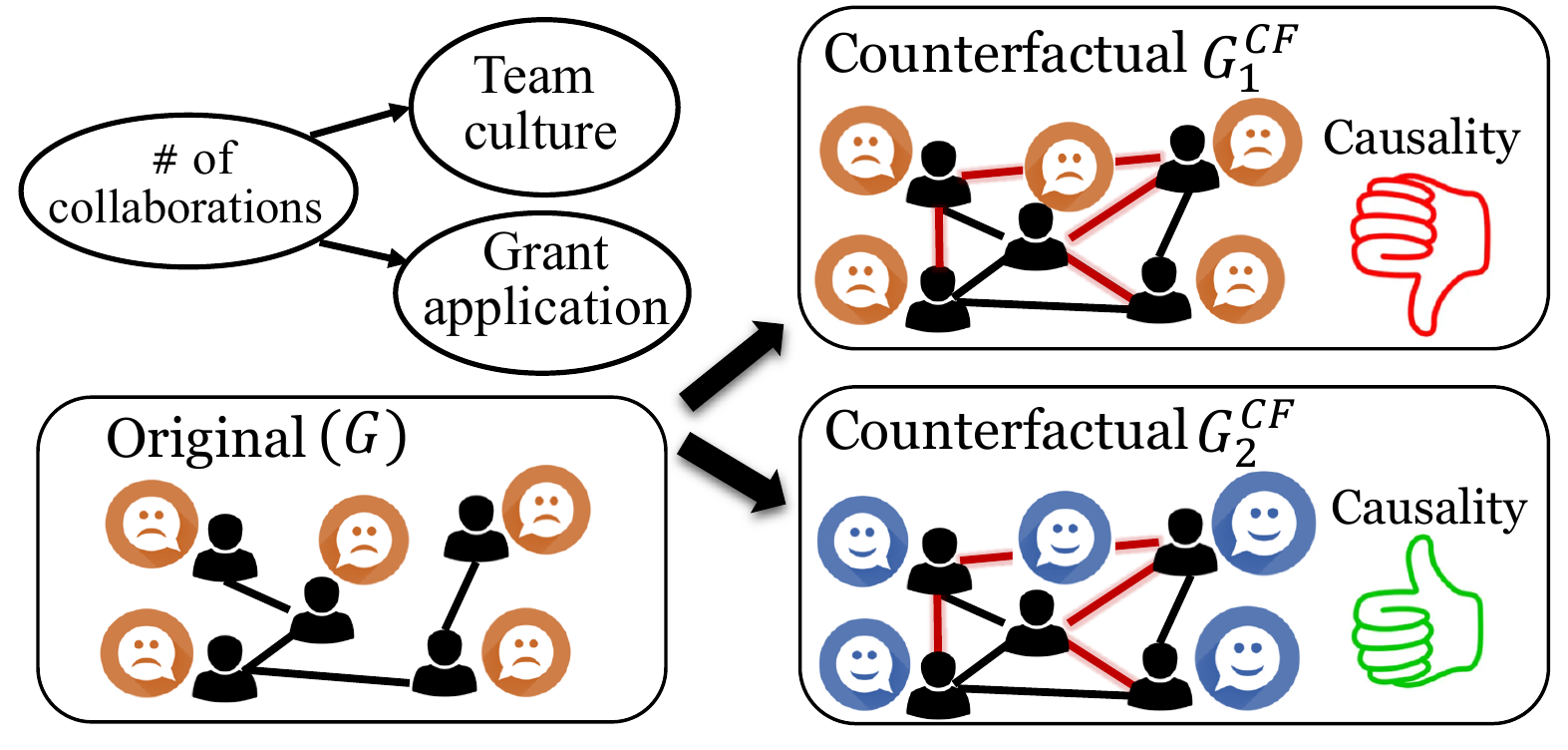} 
 %   \vspace{-.75cm}
  \end{center}
  \caption{An example of causality in CFE.} 
%\vspace{-.5cm}
\label{fig:causal}
\end{wrapfigure}

\begin{definition}
(Causality in CFE) For an explainee graph $G$, a counterfactual $G^{CF}$ satisfies causality if the change from $G$ to $G^{CF}$ is consistent with the underlying structural causal model. 
\end{definition}

\noindent\textbf{Example:} In the aforementioned grant application example, for a research team which has been rejected for an application before (as the graph $G$ in Fig.~\ref{fig:causal}), to get the next application approved, a valid counterfactual may suggest this team to improve the number of collaborations between team members. 
Based on real-world observations, we assume that an additional causal relation exists: the number of team collaborations causally affects the team culture. For example, for the same team, if more collaborations had been established, then the team culture should have been improved in terms of better member engagement and respect for diversity. %(if other variables influencing team culture do not change). 
The SCM is illustrated in Fig.~\ref{fig:causal}, where we leave out the exogenous variables for  simplicity.
%in the SCM shown in Fig.~\ref{fig:causal}, where we leave out the exogenous variables for illustration simplicity. In particular, we assume that the number of team collaborations causally affects the team culture. 
Although the team culture usually does not affect the result of grant application, the counterfactuals with team culture changed correspondingly when the number of collaborations changes are more consistent with the ground-truth SCM. $G^{CF}_2$ in Fig.~\ref{fig:causal} shows an example of such counterfactuals.
In contrast, if a counterfactual improves a team's number of collaborations alone without improving the team culture (see $G^{CF}_1$ in Fig.~\ref{fig:causal}), then it violates the causality.
%In the aforementioned job application example, improving an applicant's career networking (e.g., accumulate more social connections through work collaboration) can often lead to better outcome of job application, thus the counterfactuals for job application approval are supposed to improve the original career networking of the explainee applicant. We assume that there is a causal relation: each job applicant's career networking is improved with the increase of their work experience. Therefore, if a counterfactual improves an applicant's career networking alone without increasing his/her work experience, then it violates the causality. 
As discussed in \cite{mahajan2019preserving}, traditional CFE methods %which optimize on a single instance 
often optimize on a single instance, and are prone to perturb different features independently, thus they often fail to satisfy the causality. 

%In this work, we target at developing a generative framework which generates counterfactual explanations for graphs. Specifically, we aim to 1) facilitate optimization of the CFE generator without enumerating perturbation operations on graphs; 2) facilitate the generalization of the CFE generator on unseen graphs;
%genera counterfactuals for different graphs once after training; 
%3) promote causality in the generated counterfactuals without explicit prior knowledge of the causal model.
%\vspace{-2mm}
\section{The Proposed Framework --- \mymodel}
%\vspace{-2mm}
 
In this section, we describe a novel generative framework --- \mymodel, which addresses the problem of counterfactual explanation generation for graphs. First, we introduce its backbone \mymodel-VAE to enable optimization on graphs and generalization on unseen graph instances. This backbone is based on a graph variational auto-encoder (VAE) \cite{simonovsky2018graphvae} mechanism. %with an encoder and a decoder. The encoder maps each original graph $G=(X,A)$ into a latent space as a representation $Z$, then the decoder generates a counterfactual $G^{CF}$ based on the latent representation $Z$.
On top of \mymodel-VAE, we then promote the causality of CFEs with an auxiliary variable. 
%by improving the model identifiability.

\begin{figure*}[t]
\centering
        \includegraphics[width=.9\textwidth,height=2.2in]{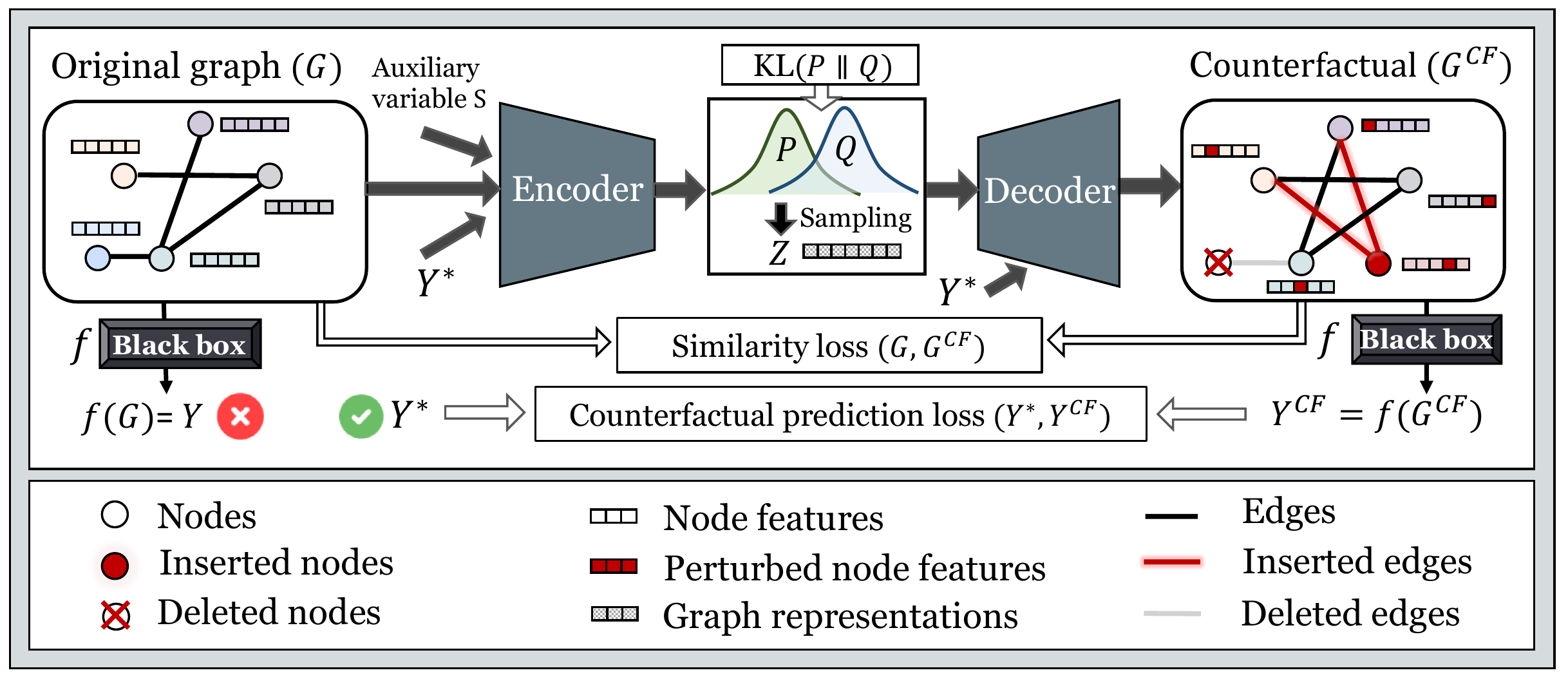}
        %\vspace{-0.1in}
        \caption{An illustration of the proposed framework \mymodel. %The framework follows a graph VAE mechanism as backbone. The encoder maps the original explainee graph and a desired label into a latent representation. Then the decoder generates  counterfactuals based on the representation. An auxiliary variable is incorporated to promote the causality of counterfactuals.
        } 
        %\vspace{-0.1in}
         \label{fig:framework}
 \end{figure*}
 
%\vspace{-2mm}
\subsection{\mymodel-VAE: Backbone of Graph Generative Counterfactual Explanations}
%\vspace{-2mm}
Different from most existing methods \cite{lucic2021cf,ying2019gnnexplainer} for CFE generation on graphs which focus on a single graph instance, \mymodel~is based on a generative backbone \mymodel-VAE~which can efficiently generate CFEs for different graphs after training, even for the graphs that do not appear in the training data. As shown in Fig.~\ref{fig:framework}, \mymodel-VAE follows a traditional graph VAE \cite{simonovsky2018graphvae} architecture with an encoder and a decoder. The encoder maps each original graph $G=(X,A)$ into a latent space as a representation $Z$, then the decoder generates a counterfactual $G^{CF}$ based on the latent representation $Z$.
Following the VAE mechanism as in  \cite{simonovsky2018graphvae,kingma2013auto} and its recent application in CFE generation for tabular data \cite{mahajan2019preserving,pawelczyk2020learning}, the optimization objective is based on
the evidence lower bound (ELBO), which is a lower bound for the log likelihood $\ln P(G^{CF}|Y^{*},G)$. Here, $P(G^{CF}|Y^{*},G)$ is the probability of the generated counterfactual $G^{CF}$ conditioned on an input explainee graph $G$ and a desired label $Y^{*}$. The ELBO for \mymodel-VAE can be derived as follows:
% \begin{equation}
%     \ln \! P(G^{CF}|Y^{*}\!,G)\!\!\ge\!\! \mathbb{E}_{Q}[\ln P(G^{CF}|Z, Y^{*}, G)] \!-\! \text{KL}(Q(Z|G, Y^{*})\|P(Z|G, Y^{*}),
% \end{equation}
\begin{equation}
    \begin{split}
    \ln P(G^{CF}|Y^{*},G)
    \ge \mathbb{E}_{Q}[\ln P(G^{CF}|Z, Y^{*}, G)] - \text{KL}(Q(Z|G, Y^{*})\|P(Z|G, Y^{*})),
\end{split}
\end{equation}
%where $Z$ is the graph representation  in the bottleneck layer for $G$, which can be considered as a set of latent variables. 
where $Q$ refers to the approximate posterior distribution  $Q(Z|G,Y^{*})$, and $\text{KL}(\cdot||\cdot)$ means the Kullback-Leibler (KL) divergence. The first term $P(G^{CF}|Z, Y^{*}, G)$ denotes the probability of the generated counterfactual conditioned on the representation $Z$ and the desired label $Y^{*}$. Due to the lack of ground-truth counterfactuals, it is hard to directly optimize this term. But inspired by \cite{mahajan2019preserving}, maximizing this term can be considered as generating valid graph counterfactuals w.r.t. the desired label $Y^{*}$, thus we replace this term with
$
 -\mathbb{E}_Q[d(G,G^{CF})+\alpha\cdot l(f(G^{CF}),Y^{*})],
$
where $d(\cdot,\cdot)$ is a similarity loss, which is a distance metric to measure the difference between $G$ and $G^{CF}$, $l(\cdot)$ is the counterfactual prediction loss to measure the difference between the predicted label $f(G^{CF})$ and the desired label $Y^{*}$. In summary, these two terms encourage the model to output counterfactuals which are similar as the input graph but elicit the desired predicted label. $\alpha$ is a hyperparameter to control the weight of the counterfactual prediction loss. Overall, the loss function of \mymodel-VAE~is: %formulated as:
\begin{equation}
\begin{split}
    \mathcal{L} = \mathbb{E}_Q[d(G,G^{CF}) +\alpha\cdot l(f(G^{CF}),Y^{*})] + \text{KL}(Q(Z|G, Y^{*})\|P(Z|G, Y^{*})).
\end{split}
\label{eq:loss}
\end{equation}

%\vspace{-1mm}
%\subsubsection{Encoder}
\noindent\textbf{Encoder.} In the encoder, the input includes node features $X$ and graph structure $A$ of the explainee graph $G$, as well as the desired label $Y^{*}$, the output is latent representation $Z$. The encoder learns the distribution $Q(Z|G,Y^{*})$. We use a Gaussian distribution  $P(Z|G,Y^{*})=\mathcal{N}(\mu_z({Y^{*})},\text{diag}(\sigma^2_z{(Y^{*}))})$ as prior, and enforce the learned distribution $Q(Z|G,Y^{*})$ to be close to the prior by minimizing their KL divergence. Here, $\mu_z({Y^{*})}$ and $\text{diag}(\sigma^2_z(Y^{*}))$ are mean and diagonal covariance of the prior distribution learned by a neural network module. $Z$ is sampled from the learned distribution $Q(Z|G,Y^{*})$ with the widely-used reparameterization trick \cite{kingma2013auto}.

%\vspace{-1mm}
%\subsubsection{Decoder}
\noindent\textbf{Decoder.} In the decoder, the input includes $Z$ and $Y^{*}$, while the output is the counterfactual $G^{CF}=(X^{CF},A^{CF})$.  Different counterfactuals can be generated for one explainee graph by sampling $Z$ from $Q(Z|G,Y^{*})$ for multiple times. The adjancency matrix is often discrete, and typically assumed to include only binary values ($A_{(i,j)}=1$ if edge from node $i$ to node $j$ exists, otherwise $A_{(i,j)}=0$). To facilitate optimization, inspired by recent graph generative models  \cite{lucic2021cf,simonovsky2018graphvae}, our decoder outputs a probabilistic adjacency matrix $\hat{A}^{CF}$ with elements in range $[0,1]$, and then generates a binary adjancency matrix $A^{CF}$ by sampling from Bernoulli distribution with probabilities in $\hat{A}^{CF}$. We calculate the similarity loss in Eq. (\ref{eq:loss}) as:
%The decoder is the key component for graph CFE generation. 
%Generally, different graph generation methods \cite{simonovsky2018graphvae,you2018graphrnn} can be used for implementation. Inspired by \cite{simonovsky2018graphvae}, we assume that the maximum number of nodes in the graph is $K$ to make the graph generation process computationally tractable (even for dense graphs). The output $G^{CF}$ includes a $K\times K$ adjacency matrix and $K \times d$ node features.
\begin{equation}
    d(G, G^{CF}) = d_A({A}, \hat{A}^{CF}) +  \beta \cdot d_X({X},X^{CF}),
\end{equation}
% any differential distance metrics?
where $d_A$ and $d_X$ are metrics to measure the distance between two graphs w.r.t. their graph structures and node features, respectively. $\beta$ controls the weight for the similarity loss w.r.t. node features. More details of model implementation are in Appendix B. %After training, \mymodel-VAE can be easily generalized to unseen graphs.
%Here, we use $\tilde{(\cdot)}$ to denote the original graph which is matched with the counterfactual. We use a matching matrix $M=\{0,1\}^{K\times n}$. $\tilde{A}=MAM^\top$, where $M_{i,j}=1$ if and only if node $i\in G^{CF}$ and $j \in G$ and $M_{i,j}=0$ otherwise. Similar as \cite{simonovsky2018graphvae}, we obtain $M$ based on each pair of $G$ and $G^{CF}$ with graph matching algorithms.

%\vspace{-2mm}
\subsection{\mymodel: Improving the Causality in Counterfactual Explanations}
%\vspace{-2mm}
To further incorporate the causality in the generated CFEs, most existing studies \cite{mahajan2019preserving,karimi2020algorithmic,karimi2021algorithmic} leverage certain prior knowledge (e.g., a given path diagram which  depicts the causal relations among variables) of the SCM. However, it is often difficult to obtain sufficient prior knowledge of the SCM in real-world data, especially for graph data. 
In this work, we do not assume the access of the prior knowledge of SCM, and only assume that the observational data is available. However, the key challenge, as shown in \cite{karimi2020algorithmic}, is that it is impossible to \textit{identify} the ground-truth SCM from observational data without additional assumptions w.r.t. the structural equations and the exogenous variables, because different SCMs may result in the same observed data distribution.
Considering that different SCMs can generate different counterfactuals, the identifiability of SCM is an obstacle of promoting causality in CFE. Fortunately, enlightened by recent progress in nonlinear independent component analysis (ICA) \cite{khemakhem2020variational,hyvarinen2019nonlinear}, we make an initial exploration to promote causality in CFE by improving the identifiability of the latent variables in our CFE generator with the help of an auxiliary observed variable. This CFE generator is denoted by \mymodel.

%\subsubsection{Improving causality}
In nonlinear independent component analysis (ICA) \cite{khemakhem2020variational,hyvarinen2019nonlinear,shimizu2006linear,monti2020causal}, it is assumed that the observed data, e.g., $X$, is generated from a smooth and invertible nonlinear transformation of independent latent variables (referred to as \textit{sources}) $Z$. Identifying the sources and the transformation are the key goals in nonlinear ICA. Similarly, traditional VAE models also assume that the observed features $X$ are generated by a set of latent variables $Z$. %and learn a generative model $p(X,Z)$ and an inference model $q(Z|X)$ to approximates the posterior $p(Z|X)$. 
However, traditional VAEs cannot be directly used for nonlinear ICA as they lack identifiability, i.e., we can find different $Z$ that lead to the same observed data distribution $p(X)$. %It has been proved that any model with unconditioned $p(Z)$ is unidentifiable. 
Recent studies \cite{khemakhem2020variational} have shown that the identifiability of VAE models can be improved with an auxiliary observed variable $S$ (e.g., a time index or class label), which enables us to use VAE for nonlinear ICA problem. As discussed in \cite{shimizu2006linear,monti2020causal}, a SCM can be considered as a nonlinear ICA model if the exogenous variables in the SCM are considered as the sources in nonlinear ICA. Similar connections can be built between the structural equations in SCM and the transformations in ICA. Such connections shed a light on improving the identifiability of the underlying SCM without much explicit prior knowledge of the SCM.

With this idea, based on the backbone \mymodel-VAE, \mymodel~%still uses a VAE mechanism to infer the exogenous variables, but 
improves the causality in counterfactuals by promoting identifiability with an observed auxiliary variable $S$.
Intuitively, we expect the graph VAE can capture the exogenous variables of the SCM in its representations $Z$, and approximate the data generation process from the exogenous variables to the observed data, which is consistent with the SCM. Here, for each graph, the auxiliary variable $S$ can provide additional information for \mymodel~to better identify the exogenous variables in the SCM, and thus can elicit counterfactuals with better causality.
To achieve this goal, following the previous work of nonlinear ICA \cite{khemakhem2020variational,hyvarinen2019nonlinear}, we make the following assumption:

\begin{assumption}
%\vspace{-1mm}
We assume that the prior on the latent variables $P(Z|S)$ is conditionally factorial. %Here $S$ is an additional observed variable.
%\vspace{-1mm}
\end{assumption}

With this assumption, the original data can be stratified by different values of $S$, and each separated data stratum can be considered to be generated by the ground-truth SCM under certain constraints (e.g., the range of values that the exogenous variables can take). When the constraints become more restricted, the space of possible SCMs that can generate the same data distribution in each data stratum shrinks. In this way, the identification of the ground-truth SCM can be easier if we 
leverage the auxiliary variable $S$.
Here, with the auxiliary variable $S$, we infer the ELBO of \mymodel: 

\noindent\textbf{Theorem 1.} The evidence lower bound (ELBO) to optimize the framework \mymodel~is:
\begin{equation}
\begin{split}
    \ln P(G^{CF}|S, Y^{*}, G) 
\ge \mathbb{E}_Q[\ln P(G^{CF}|Z, S,Y^{*}, G)] \!-\! \text{KL}(Q(Z|G, S,Y^{*})\|P(Z|G, S,Y^{*})),
\end{split}
\end{equation}
the detailed proof is shown in Appendix A. 

\noindent\textbf{Loss function of \mymodel.} Based on the above ELBO, the final loss function of \mymodel~is:
\begin{equation}
\begin{split}
    \mathcal{L} = \,\mathbb{E}_Q[d(G,G^{CF}) +\alpha\cdot l(f(G^{CF}),Y^{*})] + \text{KL}(Q(Z|G, S,Y^{*})\|P(Z|G, S,Y^{*})).
\end{split}
\label{eq: loss_final}
\end{equation}

\noindent\textbf{Encoder and Decoder.} 
The encoder takes the input $G$, $S$, and $Y^{*}$, and outputs $Z$ as the latent representation. We use a Gaussian prior $P(Z|G, S,Y^{*})=\mathcal{N}(\mu_z{(S,Y^{*})},\text{diag}(\sigma^2_Z{(S,Y^{*})})$ with its mean and diagonal covariance learned by neural network, and we encourage the learned approximate posterior $Q(Z|G, S,Y^{*})$ to approach the prior by minimizing their KL divergence.
Similar to the backbone \mymodel-VAE, the decoder takes the inputs $Z$ and $Y^{*}$ to generate one or multiple counterfactuals $G^{CF}$ for each explainee graph. More implementation details are in Appendix B.
\vspace{-3mm}
\section{Experiment}
\vspace{-3mm}
In this section, we evaluate our framework \mymodel~ with extensive experiments on both synthetic and real-world graphs. 
In particular, we answer the following research questions in our experiments:
%\begin{itemize}
\textbf{RQ1:} How does \mymodel~perform compared to state-of-the-art baselines?
\textbf{RQ2:} How do different components in \mymodel~contribute to the performance? %(Ablation studies) 
\textbf{RQ3:} How can the generated CFEs promote model explainability? %(Case studies: select some original-counterfactual pairs generated by our method and other methods) 
\textbf{RQ4:} How does \mymodel~perform  under different settings of hyperparameters? %(Sensitivity analysis)
%\end{itemize}

\vspace{-2mm}
\subsection{Baselines}
\vspace{-3mm}
We use the following baselines for comparison: 1) \textbf{Random}: For each explainee graph, it randomly perturbs %the node features and 
the graph structure for at most $T$ steps. Stop if a desired predicted label is achieved. %In each step, at most one edge can be inserted or removed. Stop the process if the perturbed graph can achieve a desired predicted label. 
2) \textbf{EG-IST}: For each explainee graph, it randomly inserts edges into it for at most $T$ steps. 3) \textbf{EG-RM}: For each explainee graph, it randomly removes edges for at most $T$ steps.  4) \textbf{GNNExplainer}: GNNExplainer \cite{ying2019gnnexplainer} is proposed to identify the most important subgraphs for prediction. We apply it for CFE generation 
by removing the important subgraphs identified by GNNExplainer. 5)  \textbf{CF-GNNExplainer}: CF-GNNExplainer \cite{lucic2021cf} is proposed for generating counterfactual ego networks in node classification tasks. %We use it for graph classification tasks by assigning the graph label to all the nodes inside the graph. 
We adapt CF-GNNExplainer for graph classification by taking the whole graph (instead of the ego network of any specific node) as input, and optimizing the model until the graph classification (instead of node classification) label has been changed to the desired one.
6) \textbf{MEG}: MEG \cite{numeroso2021meg} is a reinforcement learning based CFE generation method. % particularly designed for molecular analysis. %It enumerates the perturbation operations as the set of possible actions at each step.
In all the experiments, we set $T=150$. More details of the baseline setup can be referred in Appendix B.

\vspace{-2mm}
\subsection{Datasets}
\vspace{-2mm}
We evaluate our method on three datasets, including a synthetic dataset and two datasets with real-world graphs. %, including the number of graphs, average number of nodes and edges, the maximum number of nodes, number of classes, node feature dimension, and average node degree. Here we describe more details of these datasets: % as well as our preprocessing and simulation as follows:
\textbf{(1) Community.} This dataset contains synthetic graphs generated by the Erd\"{o}s-R\'{e}nyi (E-R) model \cite{erdHos1959random}. In this dataset, each graph consists of two $10$-node communities. The label $Y$ is determined by the average node degree in the first community (denoted by $\text{deg}_1(A)$). 
According to the causal model (in Appendix B), when $\text{deg}_1(A)$ increases (decreases), the average node degree in the second community $\text{deg}_2(A)$  should decrease (increase) correspondingly. 
We take this causal relation $\text{deg}_1(A) \rightarrow \text{deg}_2(A)$ as our causal relation of interest, and denote it as $R$. Correspondingly, we define a causal constraint for later evaluation of causality: ``$(\text{deg}_1(A^{CF})>\text{deg}_1(A))  \Rightarrow  (\text{deg}_2(A^{CF})<\text{deg}_2(A))"$ OR ``$ (\text{deg}_1(A^{CF})<\text{deg}_1(A))  \Rightarrow  (\text{deg}_2(A^{CF})>\text{deg}_2(A))"$. 
%\vspace{0.05in}
\textbf{(2) Ogbg-molhiv.} %Ogbg-molhiv is a molecular property prediction dataset. 
In this dataset, each graph stands for a molecule, where each node represents an atom, and each edge is a chemical bond. 
As the ground-truth causal model is unavailable, we simulate the label $Y$ and causal relation of interest $R$. 
% as follows:
% \textbf{Label generation:}  $Y\sim\text{Bernoulli}(\text{Sigmoid}(X_1-\text{AVG}_{x1}))$, where $X_1$ is the average value of a synthetic node feature over all nodes in each graph. This node feature is generated for each node from distribution $\text{Uniform}(0,1)$.  $\text{AVG}_{x1}$ means the average value of $X_1$ over all graphs.
% \textbf{Causality:} We also add a causal relation of interest $R$ between $X_1$ and another synthetic node feature $X_2$: $X_2=U_2+0.5X_1$. Here $U_2$ is simulated in a similar way as the Community dataset. Correspondingly, we have the following causal constraint: ``$(X_1^{CF}>X_1)\Rightarrow (X_2^{CF}>X_2)$" OR ``$(X_1^{CF}<X_1)\Rightarrow (X_2^{CF}<X_2)$".
%\vspace{0.05in}
\textbf{(3) IMDB-M.} This dataset contains movie collaboration networks from IMDB. In each graph, each node represents an actor or an actress, and each edge represents the collaboration between two actors or actresses in the same movie. 
Similarly as the above datasets, we simulate the label $Y$ and causal relation of interest $R$, and define causal constraints corresponding to $R$.
% as follows:
% \textbf{Label generation:}  $Y\sim\text{Bernoulli}(\text{Sigmoid}(\text{deg}(A)-\text{ADG}+\epsilon_y)$. $\text{deg}(A)$ is the average node degree in graph with adjacency matrix $A$. $\text{ADG}$ is the average value of $\text{deg}(A)$ over all graphs. 
% \textbf{Causality:} We also add a causal relation of interest $R$ from the average node degree to a synthetic node feature:  
% $X_1=U_1+0.5\text{deg}(A)/\text{ADG}$, where $U_1\sim \text{Uniform}[0.1S, 0.1S+0.1]$, $S\sim\text{Uniform}\{0,...,9\}$. % Here, $X_1$ is an additionally generated node feature concatenated to features of all nodes in each graph. 
% We denote the causal relation $\text{deg}(A)\rightarrow X_1$ as $R$, and define an associated causal constraint: ``$(\text{deg}(A^{CF})>\text{deg}(A))  \Rightarrow  (X_1^{CF}>X_1)"$ OR ``$ (\text{deg}(A^{CF})<\text{deg}_1(A))  \Rightarrow   (X_1^{CF}<X_1)$".
It is worth mentioning that the causal relation of interest $R$ in the three datasets covers different types of causal relations respectively: i) causal relations between variables in graph structure $A$; ii) between variables in node features $X$; iii) between variables in $A$ and in $X$. Thereby we comprehensively evaluate the performance of \mymodel~in leveraging different modalities (node features and graph structure) of graphs to fit in different types of causal relations. More details about datasets are in Appendix B. 

\vspace{-1mm}
\subsection{Evaluation Metrics}
\vspace{-1mm}
% sparsity, diversity, likelihood
%We include the following metrics for evaluation:

%\begin{itemize}
%    \item 
\noindent\textbf{Validity:} the proportion of counterfactuals which obtain the desired labels.
    % \begin{equation}
    %     \text{Validity}^c = \frac{1}{N_{CF}^c}\sum_{i\in[N_{CF}^c]}|\mathbf{1}(f(G^{CF,c}_i)=c)|,
    % \end{equation}
    % \begin{equation}
    %     \text{Validity} = \sum_c r(c)\text{Validity}^c,
    % \end{equation}
    \begin{equation}
    \small
        \text{Validity} = \frac{1}{N}\sum\nolimits_{i\in[N]}\frac{1}{N^{CF}}\sum\nolimits_{j\in[N^{CF}]}|\mathbf{1}(f(\mathbf{G}^{CF}_{(i,j)})=y^*_i)|,
    \end{equation}
    where $N$ is the number of graph instances, $N^{CF}$ is the number of counterfactuals generated for each graph. $\mathbf{G}^{CF}_{(i,j)}=(\mathbf{\mathbf{X}}^{CF}_{(i,j)}, \mathbf{A}^{CF}_{(i,j)})$ denotes the $j$-th counterfactual generated for the $i$-th graph instance. 
    %\footnote{In this paper, we use non-bold, italicized, and capitalized letters (e.g., $G$) to denote random variables; non-bold lowercase letters (e.g., $y_i$) to denote observed values of a scalar; bold capitalized letters (e.g., $\mathbf{G}$) to denote the observed values of a set/matrix.}. 
    Here, $y_i^*$ is the realization of $Y^*$ for the $i$-th graph.
    %$G^{CF,c}$ is the set of counterfactual explanations when the desired label $Y^{CF}=c$, here $|G^{CF,c}|=N_{CF}^c$, and $r(c)$ is the ratio of CFEs corresponding to the label $c$. 
    $\mathbf{1}(\cdot)$ is an indicator function which outputs $1$ when the input condition is true, otherwise it outputs $0$. 
    % \item Sparsity: how much the features/graph structure are altered.
    %\item 
    
\noindent\textbf{Proximity:} the similarity between the generated counterfactuals and the input graph. Specifically, we separately evaluate the proximity w.r.t. node features and graph structure, respectively.
    \begin{equation}
    \small
        \text{Proximity}_X \!=\! \frac{1}{N}\!\sum_{i\in[N]} \frac{1}{N^{CF}}\!\!\!\sum_{j\in[N^{CF}]} \!\!\!\text{sim}_X\!(\mathbf{X}_{(i)}, \mathbf{X}^{CF}_{(i,j)}), \,\,
        \text{Proximity}_A \!\!=\! \frac{1}{N}\!\sum_{i\in[N]} \!\frac{1}{N^{CF}}\!\!\!\sum_{j\in[N^{CF}]}\!\!\!\! \text{sim}_A(\mathbf{A}_{(i)}, \mathbf{A}^{CF}_{(i,j)}),
    \end{equation}
     where we use cosine similarity for $\text{sim}_X(\cdot)$, and  accuracy for $\text{sim}_A(\cdot)$.
   % where we use $l_2$ norm for $\text{sim}_X(\cdot)$, and use cross entropy for $\text{diff}_A(\cdot)$.
    %\item Data closeness: the distance of counterfactual explanations from data distribution. We measure data closeness by calculating the  reconstruction log-likelihood $(\log P(G^{CF}|Z,Y^{*},S))$ of the generated CFEs.
    %\item 
    
\noindent\textbf{Causality:} %we use two metrics to evaluate causality. 
    As it is difficult to obtain the true SCMs for real-world data, we % validate the causality of the CFEs w.r.t. all causal relations,
    focus on the causal relation of interest $R$ mentioned in dataset description. %calculate the log Likelihood $(\log P(R|M))$ of CFs w.r.t. a given causal edge distribution in the causal model $M$. On the other hand, we 
    Similarly as \cite{mahajan2019preserving}, we measure the causality by reporting the ratio of counterfactuals which satisfy the causal constraints corresponding to $R$. 
    %\item Correctness: we measure the correctness of counterfactuals with how much ``correct" perturbations are made in the counterfactuals to elicit the desired label. As we simulate the labels, we report the percentage of correct counterfactuals, e.g., in the Community dataset, a correct counterfactual for changing the label from $Y=0$ to $Y=1$ should increase the average node degree in the first community.
    %\item 
    
\noindent\textbf{Time:} the average time cost (seconds) of generating a counterfactual for a single graph instance.
    %\item Robustness/consistence
%\end{itemize}

\vspace{-2mm}
\subsection{Setup}
\vspace{-2mm}
We set the desired label $Y^*$ for each graph as its flipped label (e.g., if $Y=0$, then $Y^*=1$). For each graph, we generate three counterfactuals for it ($N^{CF}=3$). 
%Each dataset is randomly splitted into 60\%/20\%/20\% training/validation/test set. Unless otherwise specified, we set the hyparameters as $\alpha=5.0,\beta=10.0$. The batch size is $500$, and the representation dimension is $32$. The graph prediction models trained on all the above datasets perform well in label prediction (AUC-ROC score over 95\% and F1 score over 90\% on test data). 
Other setup details are in Appendix B.

\begin{table*}[]
\caption{The performance (mean $\pm$ standard deviation over ten repeated executions) of different methods of CFEs on graphs. The best results are in bold, and the runner-up results are underlined.}
% \vspace{-2mm}
\scriptsize
 \label{tab:baseline}
\begin{tabular}{l||l||cccccc}
\hline
\multicolumn{1}{l||}{Datasets}     &    Methods      & Validity ($\uparrow$) & Proximity$_X$($\uparrow$) & Proximity$_A$ ($\uparrow$) &  Causality ($\uparrow$) & Time ($\downarrow$)\\
\hline
% \multirow{7}{*}{Community} & Random   &    $0.116 \pm 0.047$         &       N/A         &     $0.956\pm 0.001$            &      $0.533\pm 0.165$         &  $0.021 \pm 0.008$             \\
%                           & EG-IST  &     $0.217\pm 0.047$         &      N/A          &    $\underline{0.957 \pm 0.001}$             &    $0.083 \pm 0.024$          &    $0.021\pm 0.007$           \\
%                           & EG-RMV &    $0.183\pm 0.024$          &     N/A           &      $\bm{0.960\pm 0.001}$           &   $0.017\pm 0.024$           &     $0.021\pm 0.007$          \\
\multirow{7}{*}{Community} & Random   &    $0.53 \pm 0.05$         &       N/A         &     $\underline{0.77\pm 0.02}$            &      $\underline{0.52\pm 0.06}$         &  $0.20 \pm 0.01$             \\
                           & EG-IST  &     $0.53\pm 0.05$         &      N/A          &    $0.66\pm 0.03$             &    $0.13 \pm 0.06$          &    $0.27\pm 0.03$           \\
                           & EG-RMV &    $0.55\pm 0.04$          &     N/A           &      $\bm{0.85\pm0.01}$           &   $0.03\pm 0.02$           &     $\underline{0.15\pm 0.01}$          \\
                           & GNNExplainer   &    $0.52\pm 0.06$          &      N/A          &   $0.71\pm 0.01$              &   $0.05\pm 0.00$           &    $2.87\pm 0.08$           \\
                           & CF-GNNExplainer &     $\underline{0.90 \pm 0.04}$         &    N/A            &   $0.72\pm 0.00$              &   $0.14\pm 0.02$           &   $25.14\pm 1.22$            \\
                           & MEG      &     $0.88\pm 0.04$         &   N/A            &  $0.71\pm 0.01$               &   ${0.10\pm 0.03}$           &   $27.29\pm 1.32$            \\
                           & \textbf{\mymodel(ours)}     &    $\bm{0.94\pm 0.02}$          &     $\bm{0.91\pm0.01}$           &    $0.77\pm 0.00$             &    $\bm{0.65 \pm 0.03}$         &    $\bm{0.01\pm 0.01}$           \\
\hline
% \multirow{7}{*}{Ogbg-molhiv} & Random   &  $0.516\pm 0.103$            &  N/A               & $0.988\pm 0.002$                &    $0.467\pm 0.094$          &   $0.280\pm 0.378$            \\
%                           & EG-IST   &   $0.516\pm 0.103$           &   N/A              &   $0.988\pm 0.002$              &  $0.467\pm 0.094$            &   $0.015\pm 0.007$            \\
%                           & EG-RM &     $0.516\pm 0.103$         &     N/A            &  $0.991\pm 0.002$               &   $0.467\pm 0.094$           &  $0.017\pm0.009$             \\
\multirow{7}{*}{Ogbg-molhiv} & Random   &  $0.48\pm 0.09$            &  N/A               & $0.87\pm 0.02$                &    $0.46\pm 0.1$          &   $0.17\pm 0.02$            \\
                           & EG-IST   &   $0.48\pm 0.09$           &   N/A              &   $0.83\pm 0.03$              &  $0.46\pm 0.09$            &   $0.19\pm 0.04$            \\
                           & EG-RM &     $0.483\pm 0.09$         &     N/A            &  $\bm{0.96\pm 0.01}$               &   $0.47\pm 0.09$           &  $\underline{0.17\pm0.04}$             \\
                           & GNNExplainer   &  $0.50\pm 0.01$            &  N/A  &  $0.92\pm0.00$ & $0.48\pm 0.10$            &          $2.78 \pm 0.10$                          \\
                           & CF-GNNExplainer &   $\underline{0.54\pm 0.02}$           &   N/A              &   $0.92\pm 0.01$              &   $0.49\pm 0.02$           &  $27.93\pm 1.20$             \\
                           & MEG      &    $0.49\pm 0.03$          &   N/A              &  $0.93\pm0.01$               &    $\underline{0.50\pm 0.10}$          &   $22.39\pm 2.20$            \\
                           &  \textbf{\mymodel(ours)}      &  $\bm{0.98\pm 0.01}$            &     $\bm{0.92\pm 0.02}$           &   $\underline{0.95\pm 0.01}$              &   $\bm{0.64\pm 0.02}$           &    $\bm{0.01\pm 0.00}$           \\
 \hline                          
% \multirow{7}{*}{IMDB-M} & Random   &   $0.183\pm 0.085$           &  N/A               &   $0.924\pm 0.002$              &  $0.350\pm 0.041$            &   $0.023\pm 0.007$            \\
%                           & EG-IST   &   $0.333\pm 0.118$           &    N/A             &   $0.925\pm0.003$              &    $0.450\pm 0.071$          &    $0.021\pm 0.007$           \\
%                           & EG-RM &   $0.150\pm 0.041$           &   N/A              &   $0.925\pm 0.003$              &  $0.533\pm 0.085$            &  $0.022\pm 0.006$             \\
\multirow{7}{*}{IMDB-M} & Random   &   $0.50\pm 0.04$           &  N/A               &   $0.67 \pm 0.01$              &  $0.43\pm 0.08$            &   $0.19\pm 0.01$            \\
                           & EG-IST   &   $0.56\pm 0.12$           &    N/A             &   $0.67\pm0.06$              &    $0.45\pm 0.07$          &    $\underline{0.16\pm 0.03}$           \\
                           & EG-RM &   $0.45\pm 0.11$           &   N/A              &   $\bm{0.75\pm 0.03}$              &  $\underline{0.53\pm 0.08}$            &  $0.18\pm 0.02$             \\
                           & GNNExplainer   &   $0.43\pm 0.10$           &   N/A             & $0.62\pm 0.02$                &   $0.50\pm 0.02$           &   $2.46\pm 0.50$            \\
                           & CF-GNNExplainer &   $\underline{0.95\pm 0.02}$           & N/A               &    $0.74\pm 0.02$             &   $0.51\pm 0.02$           &   $22.21\pm 1.42$            \\
                           & MEG      &   $0.90\pm 0.02$           &    N/A            &  $0.72\pm 0.02$               &    $0.51\pm 0.02$          &    $24.12\pm 1.08$           \\
                           &  \textbf{\mymodel(ours)}      &   $\bm{0.96\pm 0.01}$           &   $\bm{0.99\pm 0.00}$           &    $\underline{0.75\pm 0.01}$             &  $\bm{0.73\pm 0.01}$           &  $\bm{0.01\pm 0.00}$     \\
                           \hline
\end{tabular}
%\vspace{-2mm}
\end{table*}

%\vspace{-1mm}
\subsection{RQ1: Performance of Different Methods}
%\vspace{-1mm}
To evaluate our framework \mymodel, we compare its CFE generation performance  against the state-of-the-art baselines. From the results in Table \ref{tab:baseline}, %in which the best results are shown in bold. 
we summarize the main observations as follows: 
%\begin{itemize}
   % \item 
   1) \textbf{Validity and proximity.} Our framework \mymodel~achieves good performance in validity and proximity. This observation validates the effectiveness of our method in achieving the basic target of CFE generation. a) In validity, \mymodel~obviously outperforms all baselines on most datasets. 
   Random, EG-IST, and EG-RM perform the worst due to their random nature; GNNExplainer can only remove edges and nodes, which also limits its validity; CF-GNNExplainer and MEG perform well as their optimization is designed for CFE generation; b) In Proximity$_A$, \mymodel~outperforms all non-random baselines.
   EG-RM performs the best in Proximity$_A$ because most graphs are very sparse, thus only removing edges can change the graph relatively less than other methods. 
   As the baselines either cannot perturb node features, or their perturbation approach on node features cannot fit well in our setting, we do not compare Proximity$_X$ with them.
   % \item 
   2) \textbf{Time.} \mymodel~significantly outperforms all baselines in time efficiency. Most of the baselines generate CFEs in an iterative way, and MEG needs to enumerate all perturbations at each step. GNNExplainer and CF-GNNExplainer optimize on every single instance, which limits their generalization. All the above reasons erode their time efficiency. While in our framework, the generative mechanism enables efficient CFE generation and  generalization on unseen graphs, thus brings substantial improvement in time efficiency.
    %\item
    3) \textbf{Causality.} \mymodel~dramatically outperforms all baselines in causality. We contribute the superiority of our framework w.r.t. causality in two key factors: a) different from some baselines (e.g., GNNExplainer) optimized on each single graph, our framework can better capture the causal relations among different variables in data by leveraging the data distribution of the training set; b) our framework utilizes the auxiliary variable to better identify the underlying causal model and promote causality.
    %\item baselines
%\end{itemize}

% \begin{figure}[t]
% \centering
%   \begin{subfigure}[b]{0.32\textwidth}
%         \centering
%         \includegraphics[height=1.in]{figures/ablation_community.pdf}
%         \vspace{-1mm}
%         \caption{Community}
%     \end{subfigure}
%   \begin{subfigure}[b]{0.32\textwidth}
%         \centering
%         \includegraphics[height=1.in]{figures/ablation_oghg_molhiv.pdf}
%         \vspace{-1mm}
%         \caption{Ogbg-molhiv}
%     \end{subfigure}
%     \begin{subfigure}[b]{0.32\textwidth}
%         \centering
%         \includegraphics[height=1.in]{figures/ablation_imdb_m.pdf}
%         \vspace{-1mm}
%         \caption{IMDB-M}
%     \end{subfigure}
%     \vspace{-1mm}
%   \caption{Ablation studies.}
%   \label{fig:ablation}
%   \vspace{-2mm}
% \end{figure}

%\vspace{-2mm}
\subsection{RQ2: Ablation Study}

%\begin{wrapfigure}{l}{0.49\textwidth}
%\vspace{-.7cm}
\begin{figure}[t]
\centering
  %\begin{center}
  \begin{subfigure}[b]{0.48\textwidth}
  \centering
    \includegraphics[height=1.2in]{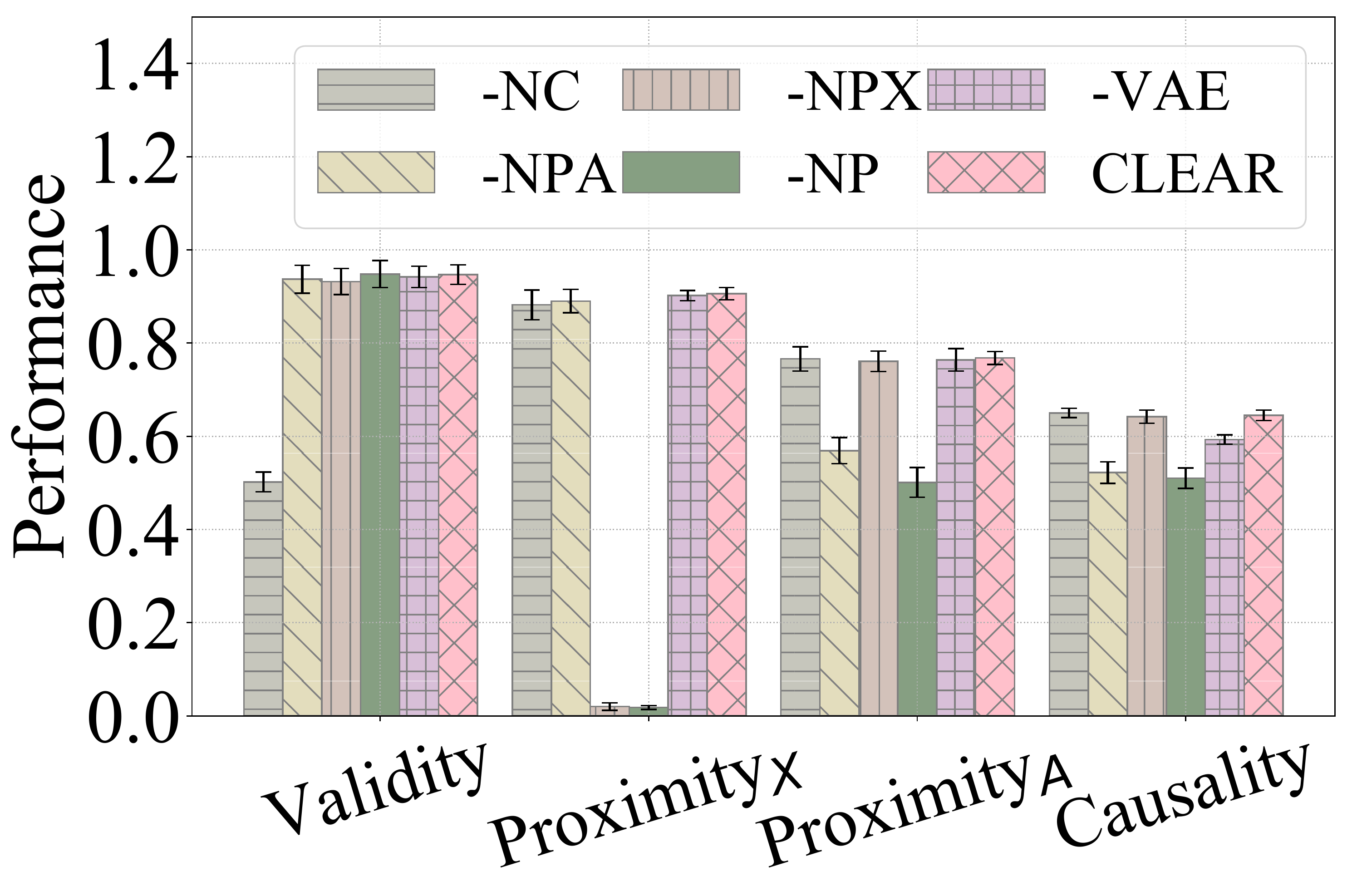} 
    \caption{Community}
    \end{subfigure}
    \begin{subfigure}[b]{0.48\textwidth}
    \centering
    \includegraphics[height=1.2in]{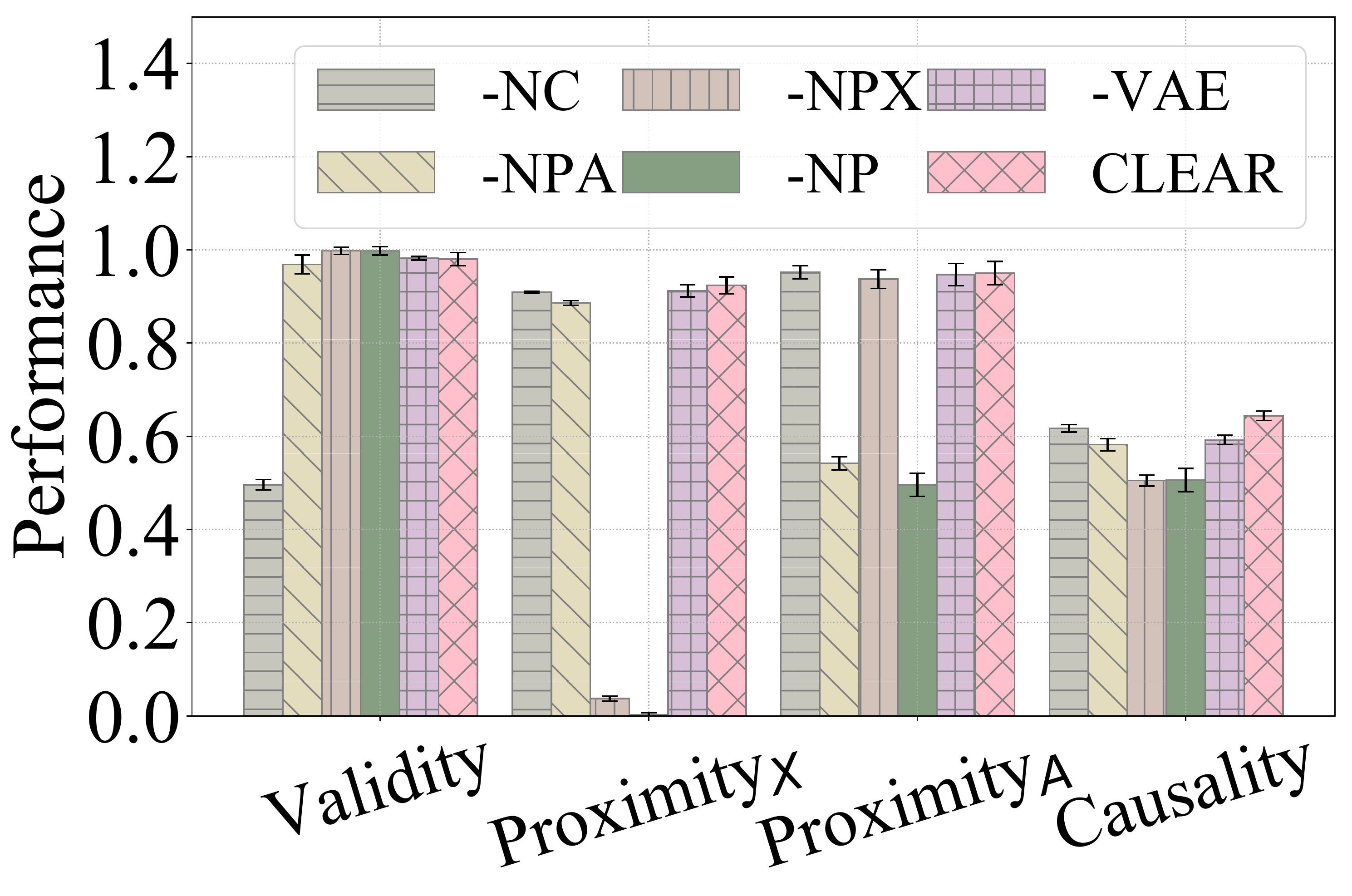} 
    \caption{Ogbg-molhiv}
    \end{subfigure}
  %  \vspace{-.6cm}
  %\end{center}
  \caption{Ablation studies.} 
%\vspace{-.5cm}
\label{fig:ablation}
%\end{wrapfigure}
\end{figure}

\begin{figure}[t]
\centering
  \begin{subfigure}[b]{0.32\textwidth}
        \centering
        \includegraphics[height=1.2in]{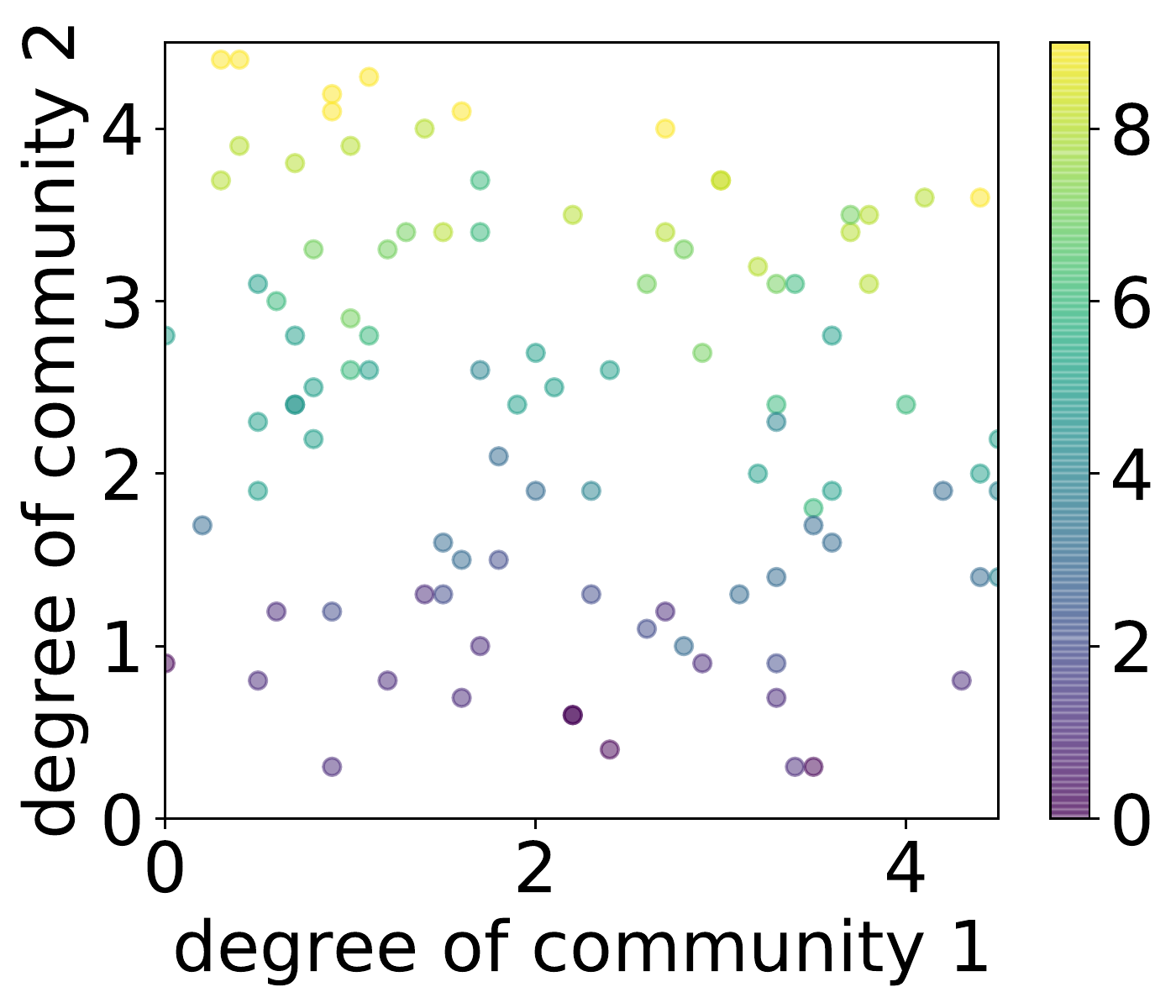}
        %\vspace{-2mm}
        \caption{Original data}
    \end{subfigure}
  \begin{subfigure}[b]{0.32\textwidth}
        \centering
        \includegraphics[height=1.2in]{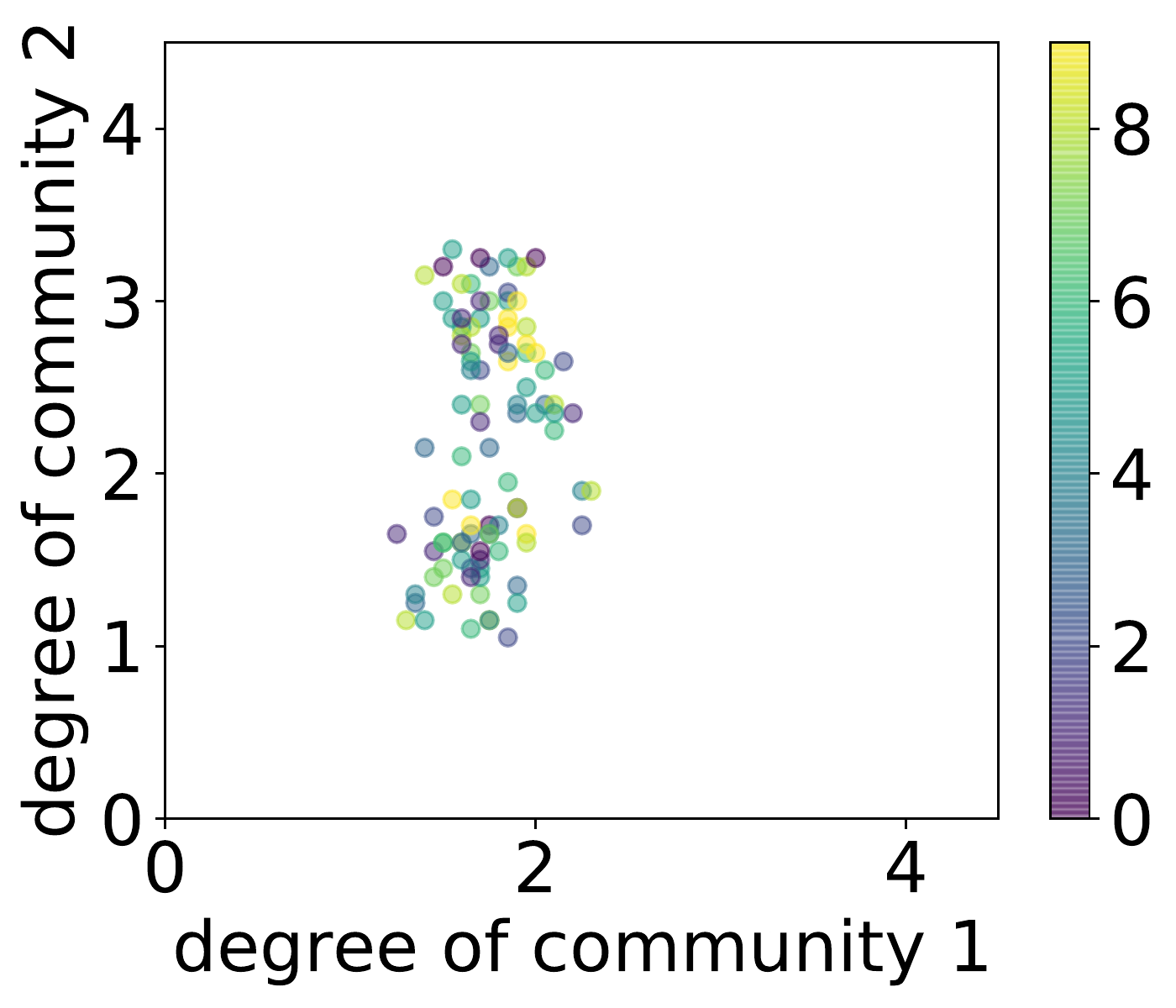}
       % \vspace{-2mm}
        \caption{CFEs from \mymodel-VAE}
    \end{subfigure}
    \begin{subfigure}[b]{0.32\textwidth}
        \centering
        \includegraphics[height=1.2in]{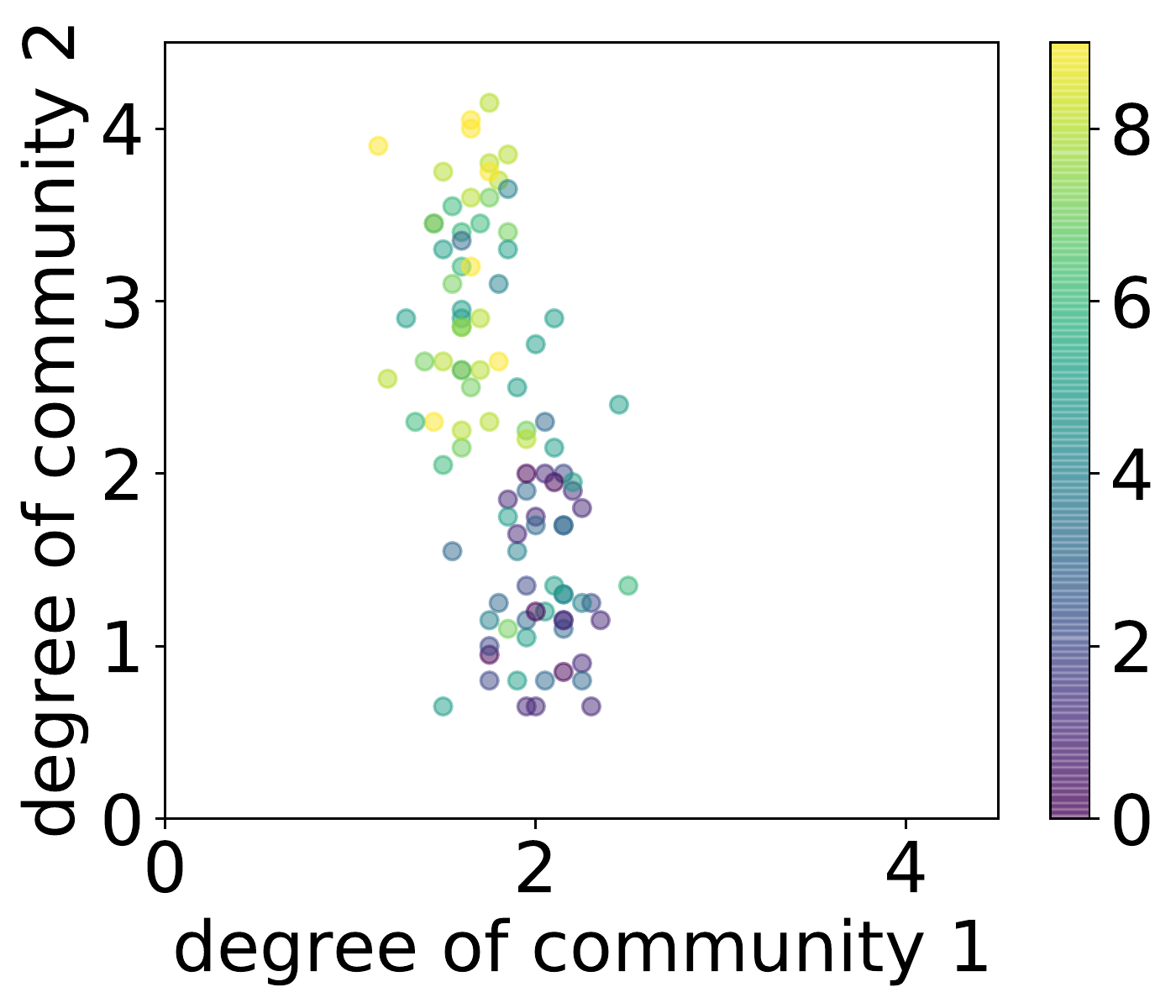}
       % \vspace{-2mm}
        \caption{CFEs from \mymodel}
    \end{subfigure}
%   \begin{subfigure}[b]{0.24\textwidth}
%         \centering
%         \includegraphics[height=1.in]{figures/case_cf_community.pdf}
%         \vspace{-2mm}
%         \caption{Case studies}
%     \end{subfigure}
   % \vspace{-1mm}
  \caption{Explainability through CFEs on Community.}
  \label{fig:case}
  %\vspace{-6mm}
\end{figure}

To evaluate the effectiveness of different components in \mymodel, we conduct ablation study with the following variants: 1) \textbf{\mymodel-NC}. In this variant, we remove the counterfactual prediction loss; 2) \textbf{\mymodel-NPA}, we remove the similarity loss w.r.t. graph structure; 3) \textbf{\mymodel-NPX}, we remove the similarity loss w.r.t. node features; 4) \textbf{\mymodel-NP}, we remove all the similarity loss; %5) \mymodel-NK, we remove the KL divergence term from the loss function. 
5) \textbf{\mymodel-VAE}, the backbone of our framework. 
As shown in Fig.~\ref{fig:ablation}, we have the following observations: 1) The validity of \mymodel-NC degrades dramatically due to the lack of counterfactual prediction loss; 2) The performance w.r.t. proximity is worse in \mymodel-NPA, \mymodel-NPX, and \mymodel-NP as the similarity loss is removed. Besides, removing the similarity loss can also hurt the performance of causality when the variables in the causal relation of interest $R$ are involved. For example, in Community, \mymodel-NPA performs much worse in causality (as $R$ in Community involves node degree in graph structure), while in Ogbg-molhiv, the performance in causality of \mymodel-NPX is eroded (as $R$ on Ogbg-molhiv involves node features); 3) The performance w.r.t. causality is impeded in \mymodel-VAE. This observation validates the effectiveness of the auxiliary variable for promoting causality. Similar observations can also be found in the ablation study on the IMDB-M dataset, which is shown in Appendix C. 
%The performance of \mymodel-NK w.r.t. causality also slightly drops, as the prior distribution learned from $S$ and $Y$

%\vspace{-2mm}
\subsection{RQ3: Explainability through CFEs}
%\vspace{-2mm}
To investigate how CFE on graphs promote model explainability, we take a closer look in the generated counterfactuals. Due to the space limit, we only show our investigation on the Community dataset. Fig.~\ref{fig:case}(a) shows the distribution of two variables: the average node degree in the first community and in the second community in the original dataset, i.e., $\text{deg}_1(A)$ and $\text{deg}_2(A)$.  
Fig.~\ref{fig:case}(b) shows the distribution of these two variables in counterfactuals generated by \mymodel-VAE. We observe that these counterfactuals are distributed close to the decision boundary, i.e., $\text{deg}_1(A)=\text{ADG}_1$, where $\text{ADG}_1$ is a constant around 2. 
This is because that the counterfactuals are enforced to change their predicted labels with perturbation as slight as possible. Fig.~\ref{fig:case}(c) shows the distribution of these two variables $\text{deg}_1(A)$ and $\text{deg}_2(A)$ in counterfactuals generated by \mymodel. 
Different colors denote different values of the auxiliary variable $S$. Notice that based on the causal model (in Appendix B), the exogenous variables are distributed in a narrow range when the value of $S$ is fixed, thus the same color also indicates similar values of exogenous variables. 
We observe that compared with the color distribution in  Fig.~\ref{fig:case}(b), the color distribution in Fig.~\ref{fig:case}(c) is  more consistent with  Fig.~\ref{fig:case}(a). %That is, compared with \mymodel-VAE, the counterfactuals generated by \mymodel~better match the original values of the exogenous variables, %($U_2$ is not supposed to be perturbed in the counterfactuals, as it is independent with the change of label), 
This indicates that compared with \mymodel-VAE, \mymodel~can better capture the values of exogenous variables, and thus the counterfactuals generated by \mymodel~are more consistent with the underlying causal model. To better illustrate the explainability provided by CFE, we further conduct case studies to compare the original graphs and their counterfactuals in Appendix C.

%To better illustrate the explainability provided by CFE, we further conduct case studies to compare the original graphs and their counterfactuals. Fig.~\ref{fig:case}(d) shows the change from original graphs to their counterfactuals w.r.t. the above two variables. We select $6$ graphs and show them in different shapes of markers. The colors denote their values of $S$ with the same colorbar in Fig.~\ref{fig:case}(a-c). We connect the pairs (original, counterfactual generated by \mymodel) with solid lines, and connect the pairs (original, counterfactual generated by \mymodel-VAE) with dashed lines. 
%We have the following observations:
%1) Compared with the input graph, the counterfactuals generated by  \mymodel-VAE and \mymodel~ both make the correct perturbations to achieve the desired label (move the variable $\text{deg}_1(A)$ across the decision boundary);
%2) The counterfactuals generated by \mymodel~better match the causality than \mymodel-VAE in two aspects: a) Qualitatively, the counterfactuals generated by \mymodel~better satisfy the causal constraints introduced in the dataset description, i.e., $\text{deg}_2(A)$ increases (decreases) when $\text{deg}_1(A)$ decreases (increases); 2) Quantitatively, the changes from original graphs to their counterfactuals fit in well with the associated structural equations $\text{deg}_1(A)\rightarrow \text{deg}_2(A)$ described in the causal model in Appendix B. %Notice that in counterfactuals, $\text{deg}_1(A)$ changes but $U_2$ should maintain its original value.
%3) Spurious correlation

%\vspace{-2mm}
\subsection{RQ4: Parameter Study}
%\vspace{-2mm}
% \begin{figure}[t]
% \centering
%   \begin{subfigure}[b]{0.24\textwidth}
%         \centering
%         \includegraphics[height=1.in]{figures/parameter_ogbg_molhiv_alpha.pdf}
%         \vspace{-2mm}
%         \caption{$\alpha$}
%     \end{subfigure}
%   \begin{subfigure}[b]{0.24\textwidth}
%         \centering
%         \includegraphics[height=1.in]{figures/parameter_ogbg_molhiv_beta.pdf}
%         \vspace{-2mm}
%         \caption{$\beta$}
%     \end{subfigure}
%     \begin{subfigure}[b]{0.24\textwidth}
%         \centering
%         \includegraphics[height=1.in]{figures/parameter_ogbg_molhiv_batchsize.pdf}
%         \vspace{-2mm}
%         \caption{Batch size}
%     \end{subfigure}
%   \begin{subfigure}[b]{0.26\textwidth}
%         \centering
%         \includegraphics[height=1.in]{figures/parameter_ogbg_molhiv_dimension.pdf}
%       \vspace{-2mm}
%         \caption{Representation dimension}
%     \end{subfigure}
%         \vspace{-2mm}
%   \caption{Parameter studies on Ogbg-molhiv.}
%   \label{fig:parameter}
%   \vspace{-3mm}
% \end{figure}

%\begin{wrapfigure}{l}{0.49\textwidth}
%\vspace{-.8cm}
\begin{figure}[t]
\centering
\begin{subfigure}[b]{0.48\textwidth}
\centering
    \includegraphics[height=1.2in]{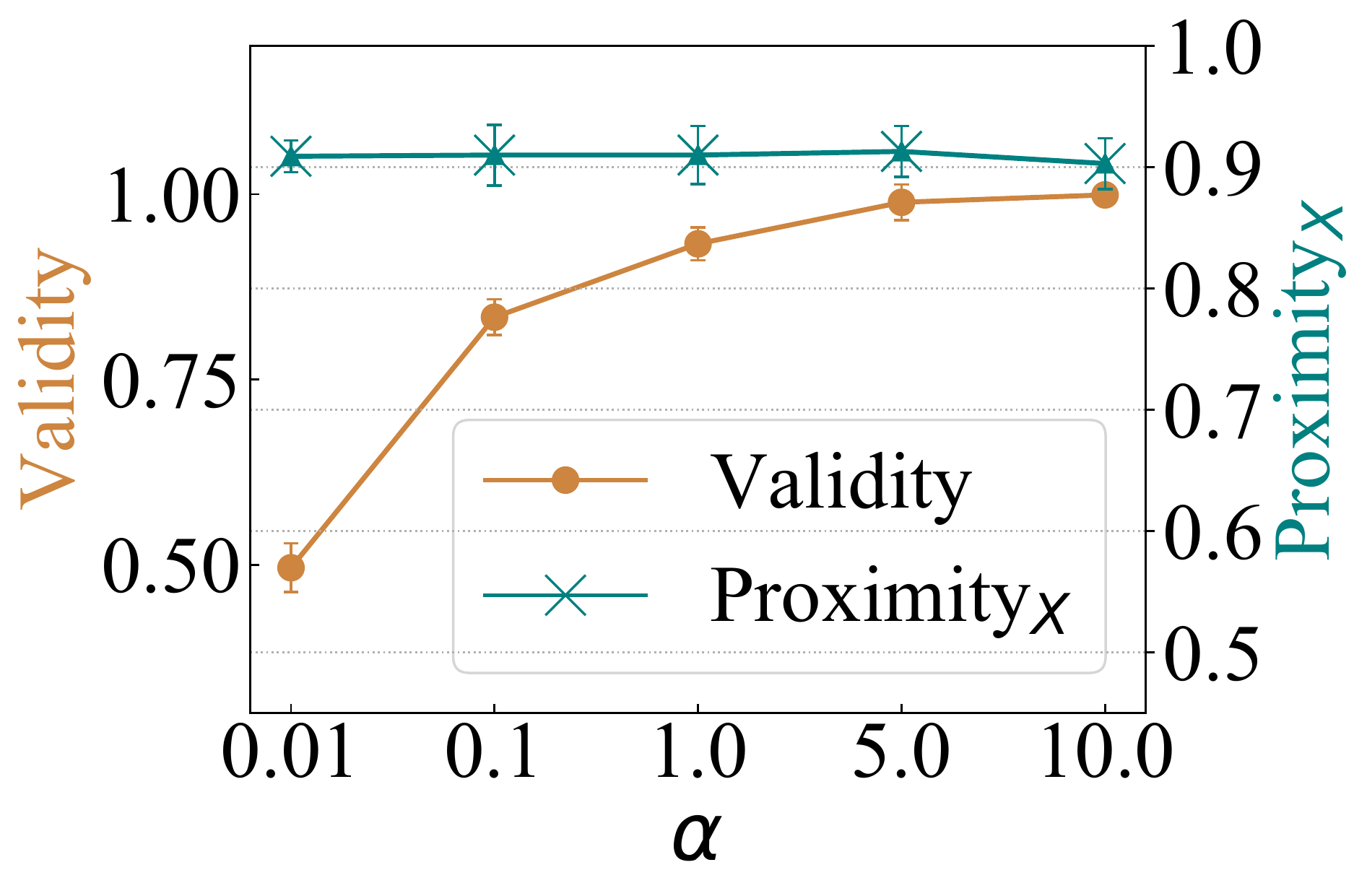} 
    \end{subfigure}
\begin{subfigure}[b]{0.48\textwidth}
\centering
    \includegraphics[height=1.2in]{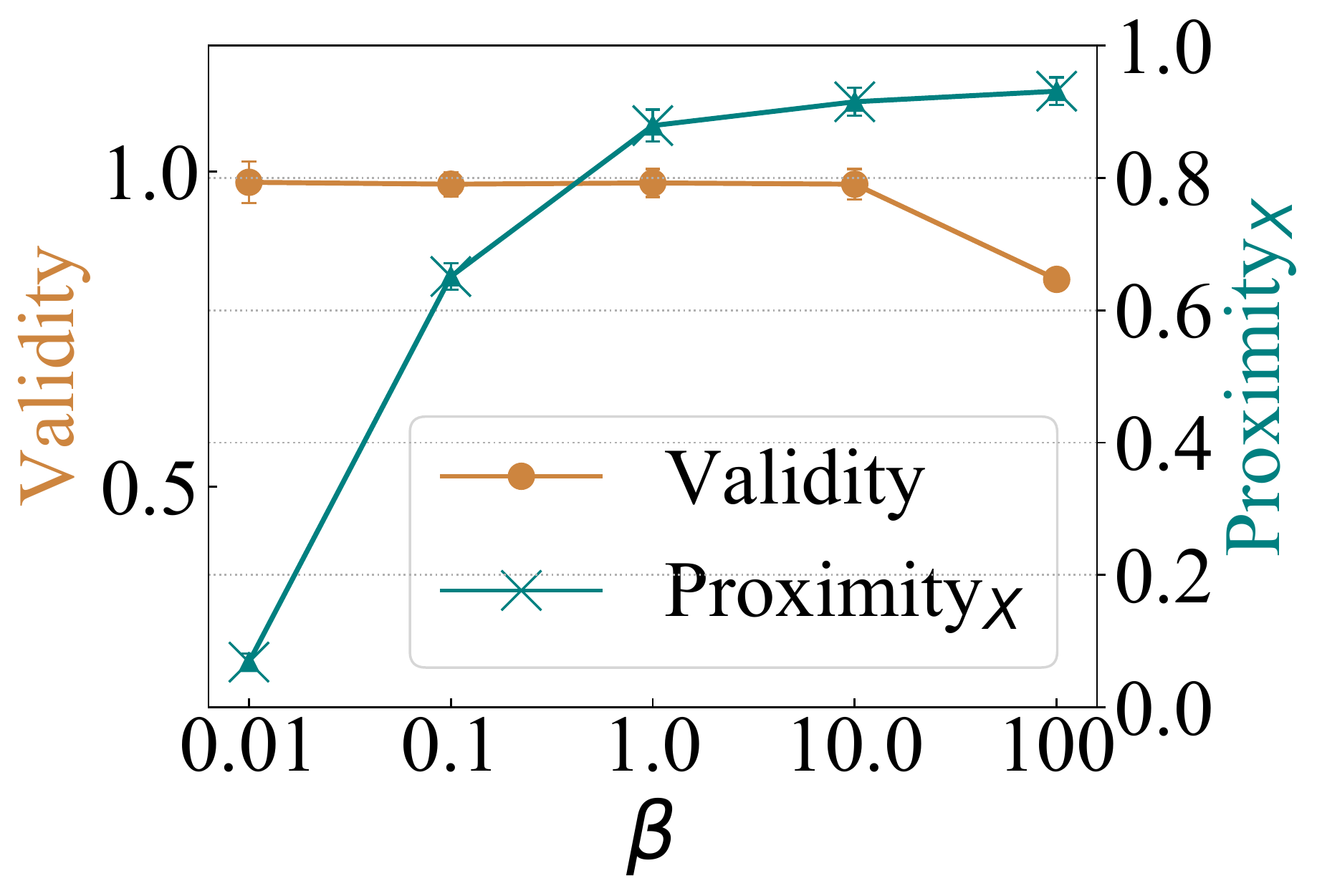} 
    \end{subfigure}
  \caption{Parameter studies on Ogbg-molhiv.} 
%\vspace{-.5cm}
\label{fig:parameter}
%\end{wrapfigure}
\end{figure}

To evaluate the robustness of our method, we test the model performance under different settings of hyperparameters. We vary $\alpha \in \{0.01, 0.1,1.0, 5.0, 10.0\}$ and $\beta\in \{0.01,0.1, 1.0, 10.0, 100\}$. %the batch size from range $\{100, 500, 1000, 2000\}$, and the representation dimension from range $\{8, 16, 32, 64\}$. 
Due to the space limit, we only show the parameter study on Ogbg-molhiv, but similar observations can be found in other datasets. As shown in Fig.~\ref{fig:parameter}, the selection of $\alpha$ and $\beta$ controls the tradeoff between different goals in the objective function. But generally speaking, the proposed framework is not very sensitive to the hyparameter setting. More studies regarding other parameters are in  Appendix C.

%\vspace{-2mm}

\vspace{-2mm}
\section{Related Work}
\vspace{-3mm}
%
% MEG vs us: we are domain-general, model-agnostic, and do not need to enumerate the action spaces.
% CF-GNN vs us: we focus on graph classification task, and can generate edge add operations, and use better difference metrics.
\noindent\textbf{Counterfactual explanations on tabular data.}
Counterfactual explanations have attracted increasing attentions \cite{wachter2017counterfactual,verma2020counterfactual,mishra2021survey,tolomei2017interpretable}. %Existing studies  \cite{wachter2017counterfactual} mostly formulate counterfactual explanation generation as an optimization problem for each specific explainee instance, and optimize this problem with an objective function by minimizing the counterfactual prediction loss w.r.t. the desired label, as well as the distance between the explainee instance and its counterfactuals.
Especially, recent works also consider more aspects in CFEs, such as actionability \cite{ustun2019actionable,poyiadzi2020face}, sparsity \cite{guidotti2018local,mothilal2020explaining,dandl2020multi}, data manifold closeness \cite{dhurandhar2018explanations,dhurandhar2019model}, diversity \cite{mothilal2020explaining,russell2019efficient}, feasibility \cite{mahajan2019preserving,poyiadzi2020face}, causality \cite{mahajan2019preserving,karimi2020algorithmic,karimi2021algorithmic}, and amortized inference \cite{mahajan2019preserving}. These methods include model-agnostic \cite{poyiadzi2020face,dandl2020multi,dhurandhar2019model} methods and model-accessible \cite{wachter2017counterfactual,tolomei2017interpretable} methods. 
Recently, a few studies \cite{mahajan2019preserving,bajaj2021robust} develop generative CFE generators based on variational autoencoder. However, most of current CFE approaches only focus on tabular or image datasets, and can not be directly grafted for the graph data.
%Some features, such as race and country of region, are immutable. The CFEs with perturbations on these features would be not actionable. 

\noindent\textbf{Counterfactual explanations on graphs.} There have been a few studies related to CFEs on graphs \cite{numeroso2021meg,lucic2021cf,bajaj2021robust,abrate2021counterfactual,ying2019gnnexplainer,yuan2020explainability}. 
%In a broader view of explaination on graphs, 
Among them, GNNExplainer \cite{ying2019gnnexplainer} identifies subgraphs which are important for graph neural network (GNN) model prediction \cite{kipf2016semi,velivckovic2017graph,kipf2016variational,jiao2020sub,wang2018graphgan,fey2019fast,hamilton2020graph}. %Although GNNExplainer is not originally a CFE method,
It can be adapted to generate counterfactuals by perturbing the identified important subgraphs. Similarly, RCExplainer \cite{bajaj2021robust} generates CFEs by removing important edges from the original graph, %RCExplainer for graphs by identifying a small subset of important edges and removing these edges in a counterfactual. RCExplainer can utilize the training data and apply the explainer to new graphs, 
but it is based on an assumption that GNN prediction model is partially accessible. CF-GNNExplainer \cite{lucic2021cf} studies counterfactual explanations for GNN models in node classification tasks. %It perturbs the graph structure for many iterations until a desired predicted label is achieved. % by removing edges from the input graph
Besides, a few  domain-specific methods \cite{numeroso2021meg,abrate2021counterfactual} are particularly designed for certain domains such as chemistry and brain diagnosis. %Among them, \cite{numeroso2021meg} proposes a reinforcement learning based method for CFE generation on molecular graphs. %The perturbations on graphs which comply to natural chemical rules are enumerated in a set of actions in each step. 
However, all the above methods are either limited in a specific setting (e.g., the prediction model is accessible), or heavily based on domain knowledge. Many general and important issues in model-agnostic CFE generation on graphs still lack exploration.

\noindent\textbf{Graph generative models.} Many efforts \cite{simonovsky2018graphvae,you2018graphrnn,grover2019graphite,ding2022data} have been made in graph generative models recently. 
Among them, GraphVAE \cite{simonovsky2018graphvae} develops a VAE-based mechanism to generate graphs from continuous embeddings. %encode and reconstruct input graphs with the representation in the bottleneck layer. 
GraphRNN \cite{you2018graphrnn} is an autoregressive generative model for graphs. It generates graphs by decomposing the graph generation process into a sequence of node and edge formations. Furthermore, a surge of domain-specific graph generative models \cite{de2018molgan,jin2018junction,preuer2018frechet} are developed with domain knowledge incorporated. 
Although different from CFE methods in their goals, graph generative models can serve as the cornerstone of our work for CFE generation on graphs. Our framework can be compatible with techniques in different graph generative models.
%In this paper, we explore the utility of graph generative methods in CFE generation on graphs.
%Graph generation can be applied in many different downstream applications, mostly in traditional tasks such as node classification and link prediction.
%Generative methods on graphs have attracted significant attention. 

%\vspace{-3mm}
\section{Conclusion}
%\vspace{-2mm}
In this paper, we study an important problem of counterfactual explanations on graphs. More specifically, we aim to facilitate the optimization, generalization, and causality in CFE generation on graphs. To address this problem, we propose a novel framework \mymodel, which uses a graph variational autoencoder mechanism to enable efficient optimization in graph data, and generalization to unseen graphs. Furthermore, we promote the causality in counterfactuals by improving the model identifiability with the help of an auxiliary observed variable. Extensive experiments are conducted to validate the superiority of the proposed framework in different aspects. In the future, more properties of the counterfactuals in graphs, such as diversity, data manifold closeness can be considered. Besides, incorporating different amount and types of prior knowledge regarding the causal models into CFE generation on graphs is also an interesting direction.

\section*{Acknowledgements}
This work is supported by the National Science Foundation under grants IIS-2006844, IIS-2144209, IIS-2223769, IIS-2106913, IIS-2008208, IIS-1955151, CNS-2154962, BCS-2228534, the JP Morgan Chase Faculty Research Award, the Cisco Faculty Research Award, and the 3 Cavaliers seed grant.
%Aidong: This work is supported in part by the US National Science Foundation under grants 2106913, 2008208, 1955151. 
Any opinions, findings, and conclusions or recommendations expressed in this material are those of the author(s) and do not necessarily reflect the views of the funding agencies.

% \small
% %% the bibliography file.

\bibliographystyle{unsrt}  % acm
\bibliography{ref}

\clearpage
\section*{Checklist}

%%% BEGIN INSTRUCTIONS %%%
% The checklist follows the references.  Please
% read the checklist guidelines carefully for information on how to answer these
% questions.  For each question, change the default \answerTODO{} to \answerYes{},
% \answerNo{}, or \answerNA{}.  You are strongly encouraged to include a {\bf
% justification to your answer}, either by referencing the appropriate section of
% your paper or providing a brief inline description.  For example:
% \begin{itemize}
%   \item Did you include the license to the code and datasets? \answerYes{See Section~\ref{gen_inst}.}
%   \item Did you include the license to the code and datasets? \answerNo{The code and the data are proprietary.}
%   \item Did you include the license to the code and datasets? \answerNA{}
% \end{itemize}
% Please do not modify the questions and only use the provided macros for your
% answers.  Note that the Checklist section does not count towards the page
% limit.  In your paper, please delete this instructions block and only keep the
% Checklist section heading above along with the questions/answers below.
%%% END INSTRUCTIONS %%%

\begin{enumerate}

\item For all authors...
\begin{enumerate}
  \item Do the main claims made in the abstract and introduction accurately reflect the paper's contributions and scope?\answerYes{}
    % \answerTODO{}
  \item Did you describe the limitations of your work?
    \answerYes{} See Appendix D.
  \item Did you discuss any potential negative societal impacts of your work? \answerYes{} See Appendix D. 
    % \answerTODO{}
  \item Have you read the ethics review guidelines and ensured that your paper conforms to them?\answerYes{}
    % \answerTODO{}
\end{enumerate}

\item If you are including theoretical results...
\begin{enumerate}
  \item Did you state the full set of assumptions of all theoretical results? \answerYes{} See Section 3.2 and Appendix A.
    % \answerTODO{}
        \item Did you include complete proofs of all theoretical results? \answerYes{} See Appendix A.
    % \answerTODO{}
\end{enumerate}

\item If you ran experiments...
\begin{enumerate}
  \item Did you include the code, data, and instructions needed to reproduce the main experimental results (either in the supplemental material or as a URL)? \answerYes{} In the submitted supplementary materials.
    % \answerTODO{}
  \item Did you specify all the training details (e.g., data splits, hyperparameters, how they were chosen)? \answerYes{} See Section 4, Appendix B and C.
    % \answerTODO{}
        \item Did you report error bars (e.g., with respect to the random seed after running experiments multiple times)? \answerYes{} All the experimental results are provided with error bars.
    % \answerTODO{}
        \item Did you include the total amount of compute and the type of resources used (e.g., type of GPUs, internal cluster, or cloud provider)? \answerYes{} See Appendix B.
    % \answerTODO{}
\end{enumerate}

\item If you are using existing assets (e.g., code, data, models) or curating/releasing new assets...
\begin{enumerate}
  \item If your work uses existing assets, did you cite the creators? \answerYes{} See Appendix B.
  \item Did you mention the license of the assets?
    \answerYes{} See Appendix B.
  \item Did you include any new assets either in the supplemental material or as a URL?
    \answerYes{} The source code of our proposed framework is included in the supplemental material. It will be released after publication.
  \item Did you discuss whether and how consent was obtained from people whose data you're using/curating?
    \answerNA{} The creators developed the adopted datasets based on either publicly available data or data with proper consent.
  \item Did you discuss whether the data you are using/curating contains personally identifiable information or offensive content? \answerNA{} We use well-known public datasets in this paper, and they do not contain personally identifiable information or offensive content.
    % \answerTODO{}
\end{enumerate}

\item If you used crowdsourcing or conducted research with human subjects...
\begin{enumerate}
  \item Did you include the full text of instructions given to participants and screenshots, if applicable?
     \answerNA{}
  \item Did you describe any potential participant risks, with links to Institutional Review Board (IRB) approvals, if applicable?
     \answerNA{}
  \item Did you include the estimated hourly wage paid to participants and the total amount spent on participant compensation?
     \answerNA{}
\end{enumerate}

\end{enumerate}

%%%%%%%%%%%%%%%%%%%%%%%%%%%%%%%%%%%%%%%%%%%%%%%%%%%%%%%%%%%%

\clearpage
% setup of baselines
\appendix
\section{Theory}
\noindent\textbf{Theorem 1.} The evidence lower bound (ELBO) to optimize the framework is:
\begin{equation}
\begin{split}
    \ln P(G^{CF}|S, Y^{*}, G) 
\ge \mathbb{E}_Q[\ln P(G^{CF}|Z, S,Y^{*}, G)] - \text{KL}(Q(Z|G, S,Y^{*})\|P(Z|G, S,Y^{*})),
\end{split}
\end{equation}
\begin{proof}
\begin{equation}
\begin{split}
    &\ln P(G^{CF}|S, Y^{*}, G) \\
= &\,\ln \int_Z P(G^{CF},Z|S, Y^{*}, G)dZ\\
= &\,\ln \int_{Z} Q(Z|G,S,Y^{*}) \frac{P(G^{CF},Z|S,Y^{*}, G)}{Q(Z|G,S,Y^{*})} dZ\\
\ge &  \int_{Z} Q(Z|G,S,Y^{*}) \ln\frac{P(G^{CF},Z|S,Y^{*}, G)}{Q(Z|G,S,Y^{*})} dZ\\
= &\, \mathbb{E}_Q[\ln\frac{P(G^{CF},Z|S,Y^{*}, G)}{Q(Z|G,S,Y^{*})}]\\
= &\, \mathbb{E}_Q[\ln\frac{P(G^{CF}|Z, S,Y^{*}, G)\cdot P(Z|G, S,Y^{*})}{Q(Z|G,S,Y^{*})}]\\
= &\, \mathbb{E}_Q[\ln{P(G^{CF}|Z, S,Y^{*}, G)}] - \mathbb{E}_Q[\ln\frac{Q(Z|G, S,Y^{*})}{P(Z|G,S,Y^{*})}]\\
= &\,\mathbb{E}_Q[\ln P(G^{CF}|Z, S,Y^{*}, G)] - \text{KL}(Q(Z|G, S,Y^{*})\|P(Z|G, S,Y^{*})).
\end{split}
\end{equation}
\end{proof}

\section{Reproducibility}
In this section, we provide more details of model implementation and experiment setup for reproducibility of the experimental results.

\subsection{Details of Model Implementation}
\subsubsection{Details of the Prediction Model}
The prediction model $f$ is implemented with a graph neural network based model. Specifically, this prediction model includes the following components:
\begin{itemize}
    \item Three layers of graph convolutional network (GCN) \cite{kipf2016semi} with learnable node masks.
    \item Two graph pooling layers with mean pooling and max pooling, respectively.
    \item A two-layer multilayer perceptron (MLP) with batch normalization and ReLU activation function.
\end{itemize}
The prediction model uses negative log likelihood loss. The representation dimension is set as $32$. We use Adam optimizer, set the learning rate as $0.001$, weight decay as $1e\!-\!5$, the training epochs as $600$, dropout rate as $0.1$, and batch size as $500$. As shown in Table \ref{tab:prediction}, we observe that the prediction model $f$ achieves high  performance of graph classification on all datasets.

\begin{table}[H]
%\small
\centering
  \caption{Performance of the prediction model on the test data of the three datasets.}
  \label{tab:prediction}
  \begin{tabular}{llll}
    \toprule
    Dataset & Community & Ogbg-molhiv & IMDB-M\\
    \midrule
    Accuracy & $0.949\pm 0.006$ & $0.897\pm 0.004$ & $0.995\pm 0.002$\\ 
    AUC-ROC  & $0.993\pm 0.002$ & $0.997\pm 0.002$ & $1.000\pm 0.001$\\
    F1-score & $0.947\pm 0.005$ & $0.906\pm 0.004$ & $0.994\pm 0.003$\\
  \bottomrule
\end{tabular}
%\vspace{-4mm}
\end{table}

\subsubsection{Details of \mymodel}
%\subsubsection{Implementation Details} 
%\label{sec:implement}
\mymodel~is designed in a general way, which can be adaptable to different graph representation learning modules and different techniques in graph generative models. Specifically, in our implementation, we apply a graph convolution \cite{kipf2016semi} based module as the encoder, and use a multilayer perceptron (MLP) as the decoder. We also use MLPs to learn the mean and covariance of the prior distribution of the latent variables. %Alternatively, other graph representation learning methods can also be applicable. 
We choose the pairwise distance using $L_2$ norm to implement $d_X(\cdot)$, and use cross entropy loss to implement $d_A(\cdot)$. We implement the counterfactual prediction loss with the negative log likelihood loss. Following \cite{simonovsky2018graphvae}, we assume that the maximum number of nodes in the graph is $k$, and use a graph matching technique to align the input graph and counterfactuals. %More details of implementation are in the Appendix B.
%In our framework \mymodel, t
The detailed implementation contains the following components:
\begin{itemize}
    \item \textbf{Prior distribution:} Two different two-layer MLPs are used to learn the mean and covariance of the prior distribution $P(Z|G,S,Y^*)$, respectively.
    \item \textbf{Encoder:} The encoder contains a single-layer graph convolutional network, a graph pooling layer with mean pooling, and two linear layers with batch normalization and ReLU activation function to learn the mean and covariance of the approximate posterior distribution $Q(Z|G,S,Y^*)$.
    \item \textbf{Decoder:} The decoder uses two three-layer MLPs to output the node features and graph structure of the counterfactual $G^{CF}$, respectively. These MLPs use batch normalization, and take ReLU as activation function in the middle layers. At the last layer of decoder, the MLP which generates the graph structure uses Sigmoid as activation function to output a probabilistic adjancency matrix $\hat{A}^{CF}$ with elements in range $[0, 1]$.
\end{itemize}

Inspired by \cite{simonovsky2018graphvae}, we use a graph matching technique  to align the input graph and counterfactuals. Specifically, we learn a graph matching matrix $M=\{0,1\}^{k\times n}$ to match the generated counterfactual with the original explainee graph. Here, $n$ is the number of nodes in the original graph, and $M_{(i,j)}=1$ if and only if node $i$ is in $G^{CF}$ and node $j$ is in $G$, and $M_{(i,j)}=0$ otherwise. 

\subsection{Details of Experiment Setup}
\subsubsection{Baseline Settings}
Here we introduce more details of baseline setting:
\begin{itemize}
    \item \textbf{Random}: For each explainee graph, it randomly perturbs the graph structure for at most $T=150$ steps. In each step, at most one edge can be inserted or removed. We stop the process if the perturbed graph can achieve a desired predicted label. 
    \item \textbf{GNNExplainer:} For each graph, GNNExplainer \cite{ying2019gnnexplainer} outputs an edge mask which estimates the importance of different edges in model prediction. In CFE generation, we set a threshold $0.5$ and remove edges with edge mask weight smaller than the threshold. Although GNNExplainer can also identify important node features in a similar way, when we apply GNNExplainer for CFE generation, the perturbation on node features cannot be designed as straightforwardly as the perturbation on graph structure, thus we did not involve perturbation on node features in GNNExplainer.
    \item \textbf{CF-GNNExplainer:} CF-GNNExplainer \cite{lucic2021cf} is originally proposed for node classification tasks, and it only focuses on the perturbations on the graph structure. Originally, for each explainee node, it takes its neighborhood subgraph as input. To apply it on graph classification tasks, we use the graph instance as the neighborhood subgraph, and assign the graph label as the label for all nodes in the graph. We set the number of iterations to generate counterfactuals for each graph as $150$.
    \item \textbf{MEG:} MEG \cite{numeroso2021meg} is specifically proposed for molecular prediction tasks. This model explicitly incorporates domain knowledge in chemistry. The CFE generator is developed based on reinforcement learning, and it designs the reward based on the prediction on the counterfactual, as well as the similarity between the original graph and the counterfactual. In each step, MEG enumerates all possible perturbations (e.g., adding an atom) which are valid w.r.t. chemistry rules to form an action set. We apply it to general graphs by removing the constraints of domain knowledge, and enumerating the perturbations as: 1) adding or removing a node; 2) adding or removing an edge; 3) staying the same. We set the number of action steps as $150$.
\end{itemize}

\subsubsection{Datasets}

\begin{table}[t]
%\small
\centering
  \caption{Detailed statistics of the datasets.}
  \label{tab:datasets}
 %\vspace{-3mm}
  \label{tab:dataset}
  \begin{tabular}{llll}
    \toprule
    Dataset & Community & Ogbg-molhiv & IMDB-M\\
    \midrule
    \# of graphs & $10,000$ & $31,957$ & $1,160$\\ 
    Avg \# of nodes & $20$ & $20.8$ & $9.4$\\ 
    Avg \# of edges & $45.0$     & $22.4$ & $32.8$\\ 
    Max \# of nodes & $20$ & $30$ & $15$\\
    \# of classes & $2$ & $2$& $2$\\
    Feature dimension & $16$ & $11$& $2$ \\
    Avg node degree & $2.24$& $1.07$ & $3.4$\\
  \bottomrule
\end{tabular}
%\vspace{-4mm}
\end{table}

For each dataset, we filter out the graphs with the number of nodes larger than a threshold $k$. The setting of $k$ (i.e., max \# of nodes) can be found in Table \ref{tab:dataset}. As some of the baselines need to be optimized separately for each graph, we compare the performance of all methods on a small set of test data with $20$ graphs for evaluation in RQ1. For other RQs, we evaluate our framework on the whole test data.

\noindent\textbf{1. Community.} We first generate a synthetic dataset in which we can fully control the data generation process. In this dataset, each graph consists of two $10$-node communities generated using the Erd\"{o}s-R\'{e}nyi (E-R) model \cite{erdHos1959random} with edge rate $p_1$ and $p_2$,  respectively. 
Specifically, we simulate the data with the following causal model: %we generate the additional observed variable $S$, adjacency matrix $A$, node features $X$, and labels $Y$ as follows: 
$$
    S \sim \text{Uniform}(\{0,...,9\}),\,\, p_1 = U_1\sim \text{Uniform}([0,1]),
$$
$$
U_2 \!\sim\! \text{Uniform}([\delta S+b, \delta(S+1)+b]), \, p_2 \!=\! \max\{0, \min\{1, -0.15p_1+ U_2\}\!\},
$$
\begin{equation}
X\sim\mathcal{N}(0,I),\,\,Y\sim\text{Bernoulli}(\text{Sigmoid}(\text{deg}_1(A) - \text{ADG}_1 +\epsilon_y)).
\end{equation}
    
$U_1$ and $U_2$ are two exogenous variables associated with $p_1$ and $p_2$, respectively. Notice that $p_2$ is determined by $p_1$ and $U_2$. Here, the auxiliary variable $S$ provides help to infer the value of exogenous variables (specifically, $U_2$ in this case). We set $\delta=0.085,b=0.15$. $p_1$ and $p_2$ thereby generate the graph structure inside the two communities, respectively. We also randomly add few edges between these two communities. The edges connecting two communties are randomly generated with an edge rate of $0.05$. In this way, the adjacency matrix $A$ of each graph is simulated. 
$\text{deg}_1(A)$ (determined by $p_1$) denotes the average node degree in the first community of each graph $A$. 
\textbf{Label generation:} The label $Y$ is determined by $\text{deg}_1(A)$ together with a Gaussian noise $\epsilon_y\sim\mathcal{N}(0,0.01^2)$. $\text{ADG}_1$ is a constant, which is the average value of $\text{deg}_1(A)$ over all graphs.
\textbf{Causality:} To elicit a different predicted label, $\text{deg}_1(A)$ in the counterfactual is supposed to be perturbed, while other variables can remain the same. But considering that with the above causal model, when $\text{deg}_1(A)$ increases (decreases), the average node degree in the second community $\text{deg}_2(A)$ (determined by $p_2$) should decrease (increase) correspondingly. We take this causal relation $\text{deg}_1(A) \rightarrow \text{deg}_2(A)$ as our causal relation of interest, and denote it as $R$. Correspondingly, we define a causal constraint for evaluation of causality: ``$(\text{deg}_1(A^{CF})>\text{deg}_1(A))  \Rightarrow  (\text{deg}_2(A^{CF})<\text{deg}_2(A))"$ OR ``$ (\text{deg}_1(A^{CF})<\text{deg}_1(A))  \Rightarrow  (\text{deg}_2(A^{CF})>\text{deg}_2(A))"$. 

%. $p\sim \text{Uniform}(0.1,0.6)$ when $U=0$, and $p\sim \text{Uniform}(0.3,0.8)$ when $U=1$. $Y=f(A,X)$. $X\sim\mathcal{N}(0,I)$, $I$ is an identity matrix. We add a causal relation $R$:  $X[0]=X[0]+\text{deg}(A)$ to adjust the data.

\vspace{0.05in}
\noindent\textbf{2. Ogbg-molhiv.} %Ogbg-molhiv is a molecular property prediction dataset. 
Ogbg-molhiv is adopted from the MoleculeNet \cite{wu2018moleculenet} datasets. All molecules are preprocessed with RDKit \cite{landrum2006rdkit}. The original node features are 9-dimensional, containing atom features such as atomic number, formal charge and chirality. 
In this dataset, each graph stands for a molecule, where each node represents an atom, and each edge is a chemical bond. 
As the ground-truth causal model is unavailable, we simulate the label and causal relation of interest as follows:
\textbf{Label generation:}  $Y\sim\text{Bernoulli}(\text{Sigmoid}(X_1-\text{AVG}_{x1}))$, where $X_1$ is the average value of a synthetic node feature over all nodes in each graph. This node feature is generated for each node from distribution $\text{Uniform}(0,1)$.  $\text{AVG}_{x1}$ means the average value of $X_1$ over all graphs.
\textbf{Causality:} We also add a causal relation of interest $R$ between $X_1$ and another synthetic node feature $X_2$: $X_2=U_2+0.5X_1$. Here $U_2$ is simulated in a similar way as the Community dataset. Correspondingly, we have the following causal constraint: ``$(X_1^{CF}>X_1)\Rightarrow (X_2^{CF}>X_2)"$ OR ``$(X_1^{CF}<X_1)\Rightarrow (X_2^{CF}<X_2)"$.

\vspace{0.05in}
\noindent\textbf{3. IMDB-M.} This dataset contains movie collaboration networks from IMDB. In each graph, each node represents an actor or an actress, and each edge represents the collaboration between two actors or actresses in the same movie. Similarly as the above datasets, we simulate the label and causal relation of interest as follows:
\textbf{Label generation:}  $Y\sim\text{Bernoulli}(\text{Sigmoid}(\text{deg}(A)-\text{ADG}+\epsilon_y)$. $\text{deg}(A)$ is the average node degree in graph with adjacency matrix $A$. $\text{ADG}$ is the average value of $\text{deg}(A)$ over all graphs. 
\textbf{Causality:} We also add a causal relation of interest $R$ from the average node degree to a synthetic node feature:  
$X_1=U_1+0.5\text{deg}(A)/\text{ADG}$, where $U_1\sim \text{Uniform}[0.1S, 0.1S+0.1]$, $S\sim\text{Uniform}\{0,...,9\}$. % Here, $X_1$ is an additionally generated node feature concatenated to features of all nodes in each graph. 
We denote the causal relation $\text{deg}(A)\rightarrow X_1$ as $R$, and define an associated causal constraint: ``$(\text{deg}(A^{CF})>\text{deg}(A))  \Rightarrow  (X_1^{CF}>X_1)"$ OR ``$ (\text{deg}(A^{CF})<\text{deg}_1(A))  \Rightarrow   (X_1^{CF}<X_1)"$.

\subsubsection{Experiment Settings}
All the experiments are conducted in the following environment:
\begin{itemize}
    \item Python 3.6
    \item Pytorch 1.10.1
    \item Pytorch-geometric 1.7.0
    \item Scikit-learn	1.0.1
    \item Scipy	1.6.2
    \item Networkx	2.5.1
    \item Numpy	1.19.2
    \item Cuda 10.1
\end{itemize}
In all the experiments of counterfactual explanation, each dataset is randomly split into 60\%/20\%/20\% training/validation/test set. Unless otherwise specified, we set the hyperparameters as $\alpha=5.0$ and $\beta=10.0$. The batch size is $500$, and the representation dimension is $32$. The graph prediction models trained on all the above datasets perform well in label prediction (AUC-ROC score over 95\% and F1 score over 90\% on test data). 
We use NetworkX \cite{hagberg2008exploring} to generate synthetic graphs. In our CFE generator \mymodel, the learning rate is 0.001, the number of epochs is 1,000. All the experimental results are averaged over ten repeated executions. The implementation is based on Pytorch. We use the Adam optimizer for model optimization.

\section{More Experimental Results}
\subsection{Ablation Study}
Fig.~\ref{fig:ablation_imdb} shows the results of ablation studies on the IMDB-M dataset. The observations are generally consistent with the observations on other two datasets as described in Section 4.6.

% \begin{figure}[t]
% \centering
%   \begin{subfigure}[b]{0.43\textwidth}
%         \centering
%         \includegraphics[height=1.2in]{figures/ablation_imdb_m.pdf}
%     \end{subfigure}
%     %\vspace{-2mm}
%   \caption{Ablation studies on the IMDB-M dataset.}
%   \label{fig:ablation_imdb}
%   %\vspace{-3mm}
% \end{figure}

% \subsection{Case Study}
% \begin{wrapfigure}{l}{0.46\textwidth}
%   \begin{center}
%     \includegraphics[width=0.46\textwidth]{figures/case_cf_community.pdf} 
%   \end{center}
%   \caption{Case studies.} 
% \label{fig:case}
% \end{wrapfigure}

\subsection{Case Study}
\begin{figure*}[t]
%\vspace{-0.1in}
  \setlength{\belowcaptionskip}{-0.25cm}
  \begin{minipage}[t]{0.5\linewidth}
   \centering
   \includegraphics[width=0.65\linewidth,height=1.3in]{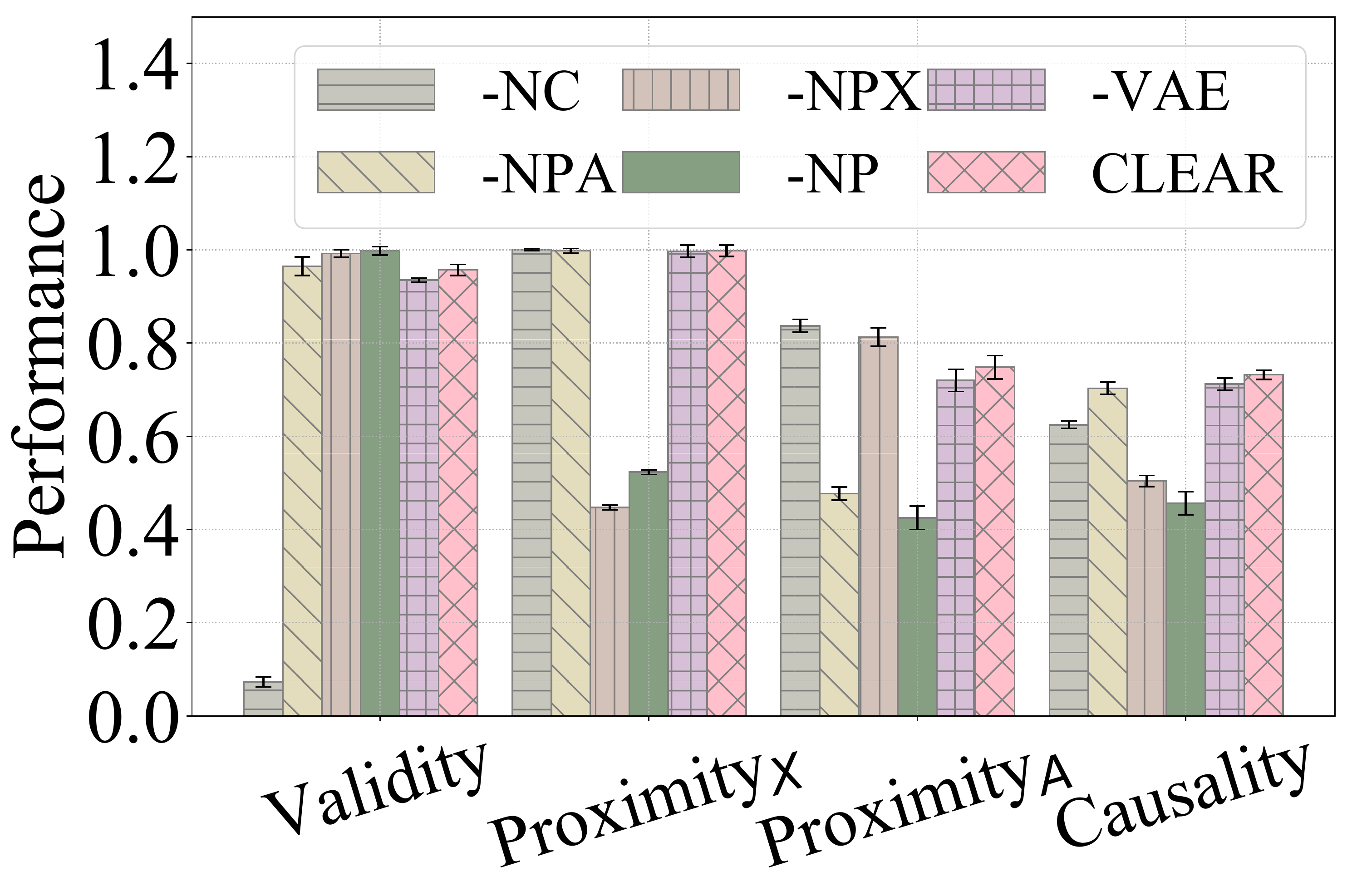}
   \captionsetup{font={footnotesize}}
   \captionsetup{justification=centering}
   %\vspace{-2mm}
   \caption{Ablation studies on the IMDB-M dataset.}
   \label{fig:ablation_imdb}
  \end{minipage}
  \hspace{0.14cm}
  \begin{minipage}[t]{0.5\linewidth}
   \centering
   \includegraphics[width=0.65\linewidth,height=1.3in]{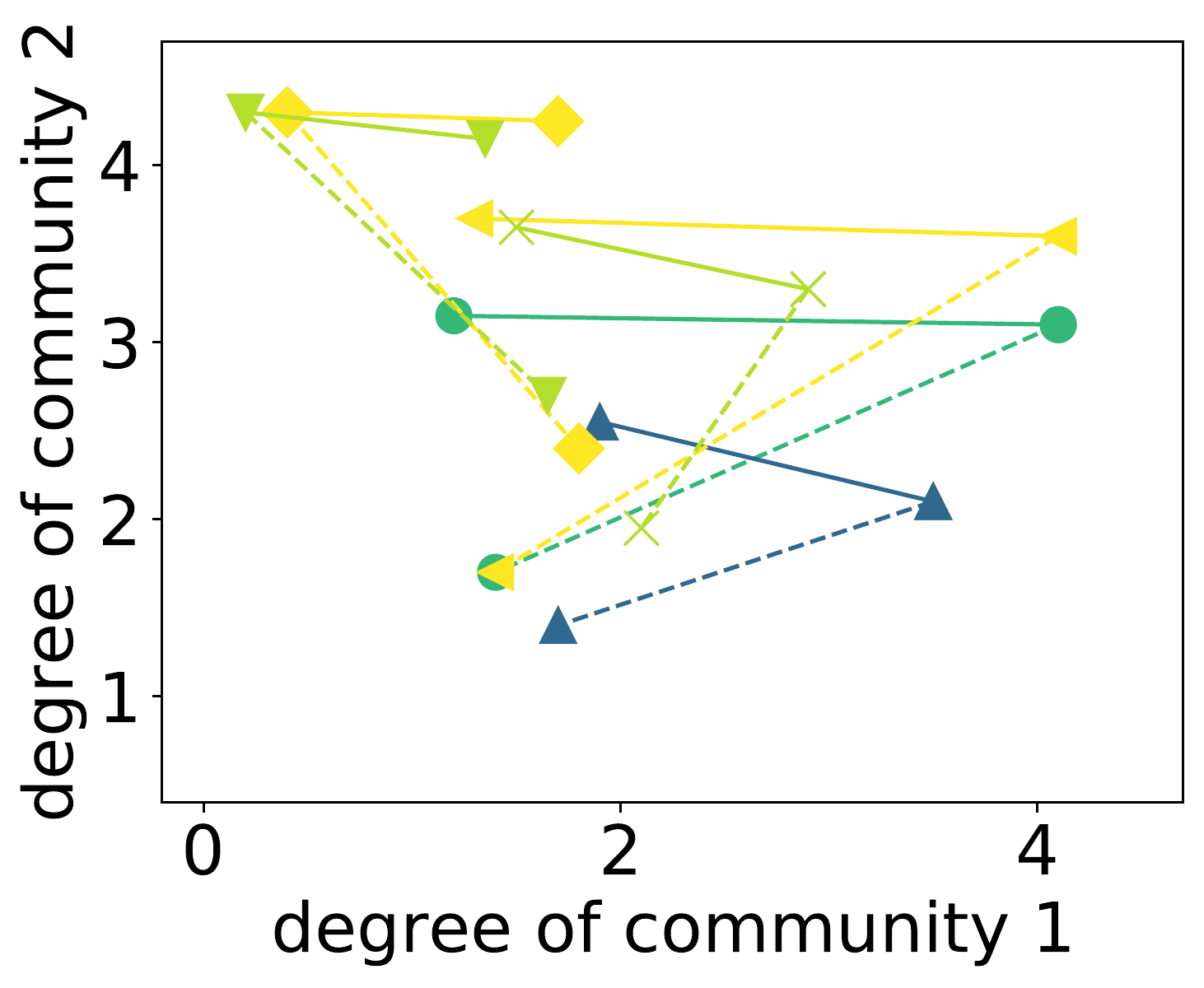}
 \captionsetup{font={footnotesize}}
   \captionsetup{justification=centering}
   %\vspace{-2mm}
   \caption{Case study.}
   \label{fig:case_pair}
  \end{minipage}
  %\hspace{0.05cm}
 % \vspace{-3mm}
\end{figure*}

To better illustrate the explainability provided by CFE, we further conduct case studies to compare the original graphs and their counterfactuals. In the Community dataset, Fig.~\ref{fig:case_pair} shows the change from original graphs to their counterfactuals w.r.t. the average node degree in the first community and in the second community, i.e., $\text{deg}_1(A)$ and $\text{deg}_2(A)$. Here, Fig. \ref{fig:case_pair} has the same x-axis and y-axis as Fig.~\ref{fig:case}. In Fig.~\ref{fig:case_pair}, we randomly select $6$ graphs and show them in different shapes of markers. The colors denote their values of $S$ with the same colorbar in Fig.~\ref{fig:case}(a-c). In Fig.~\ref{fig:case_pair}, we connect the pairs (original, counterfactual generated by \mymodel) with solid lines, and connect the pairs (original, counterfactual generated by \mymodel-VAE) with dashed lines. 
We have the following observations:
1) Compared with the input graph, the counterfactuals generated by  \mymodel-VAE and \mymodel~both make the correct perturbations to achieve the desired label (moving the variable $\text{deg}_1(A)$ across the decision boundary at around $\text{deg}_1(A)=2$);
2) The counterfactuals generated by \mymodel~better match the causality than \mymodel-VAE in two aspects: a) Qualitatively, the counterfactuals generated by \mymodel~better satisfy the causal constraints introduced in the dataset description, i.e., $\text{deg}_2(A)$ increases (decreases) when $\text{deg}_1(A)$ decreases (increases); 2) Quantitatively, the changes from original graphs to their counterfactuals fit in well with the associated structural equations $(\text{deg}_1(A), U_2)\rightarrow \text{deg}_2(A)$. Notice that in counterfactuals, $\text{deg}_1(A)$ changes but $U_2$ is supposed to maintain its original value. 

% \begin{figure}[t]
% \centering
%   \begin{subfigure}[b]{0.5\textwidth}
%         \centering
%         \includegraphics[height=1.3in]{figures/case_cf_community.pdf}
%     \end{subfigure}
%     %\vspace{-2mm}
%  \caption{Case study.}
%   \label{fig:case_pair}
%   %\vspace{-3mm}
% \end{figure}

\begin{figure}[t]
\centering
    \begin{subfigure}[b]{0.47\textwidth}
        \centering
        \includegraphics[height=1.3in]{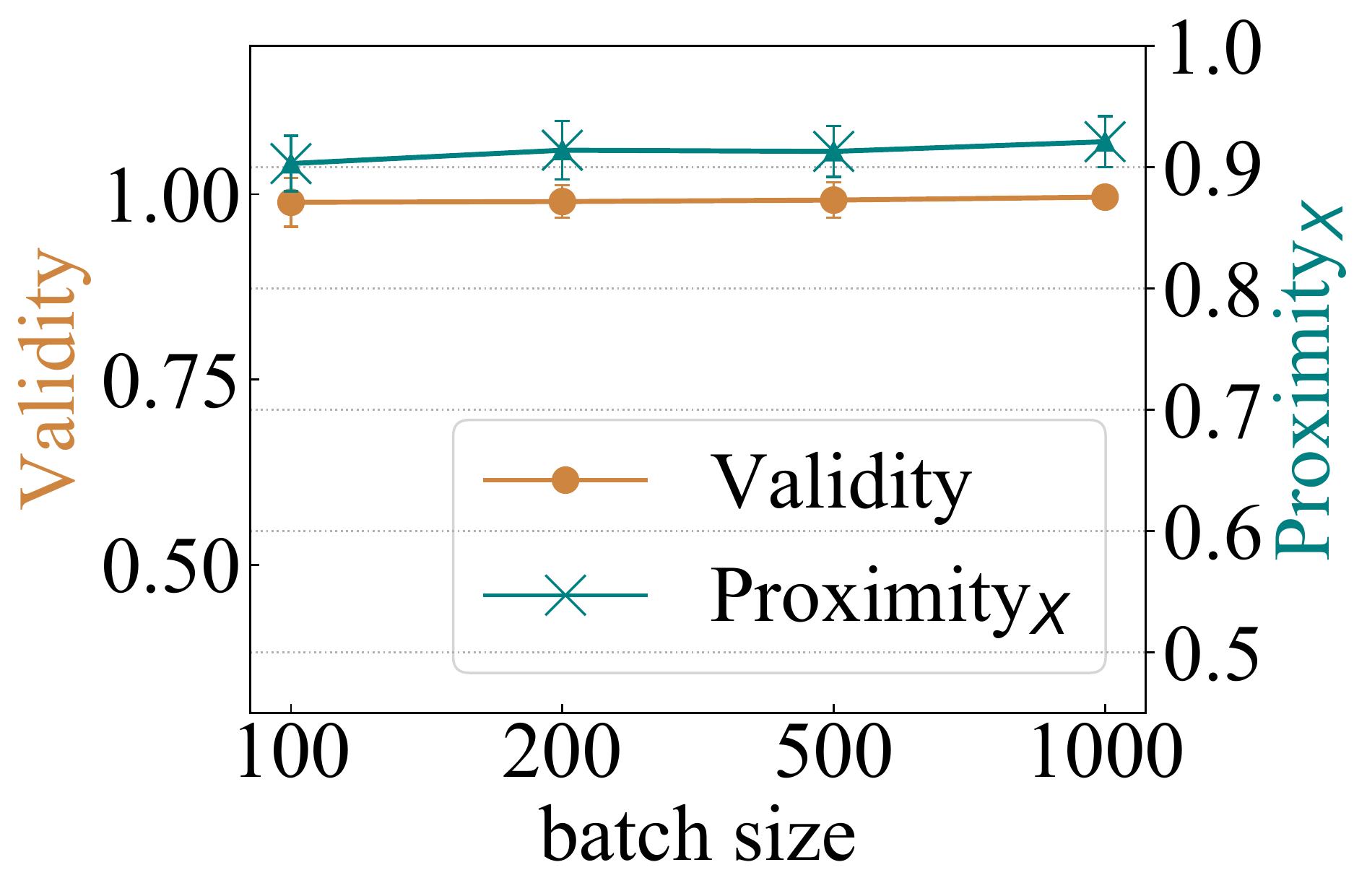}
        \vspace{-2mm}
        \caption{Batch size}
    \end{subfigure}
  \begin{subfigure}[b]{0.47\textwidth}
        \centering
        \includegraphics[height=1.3in]{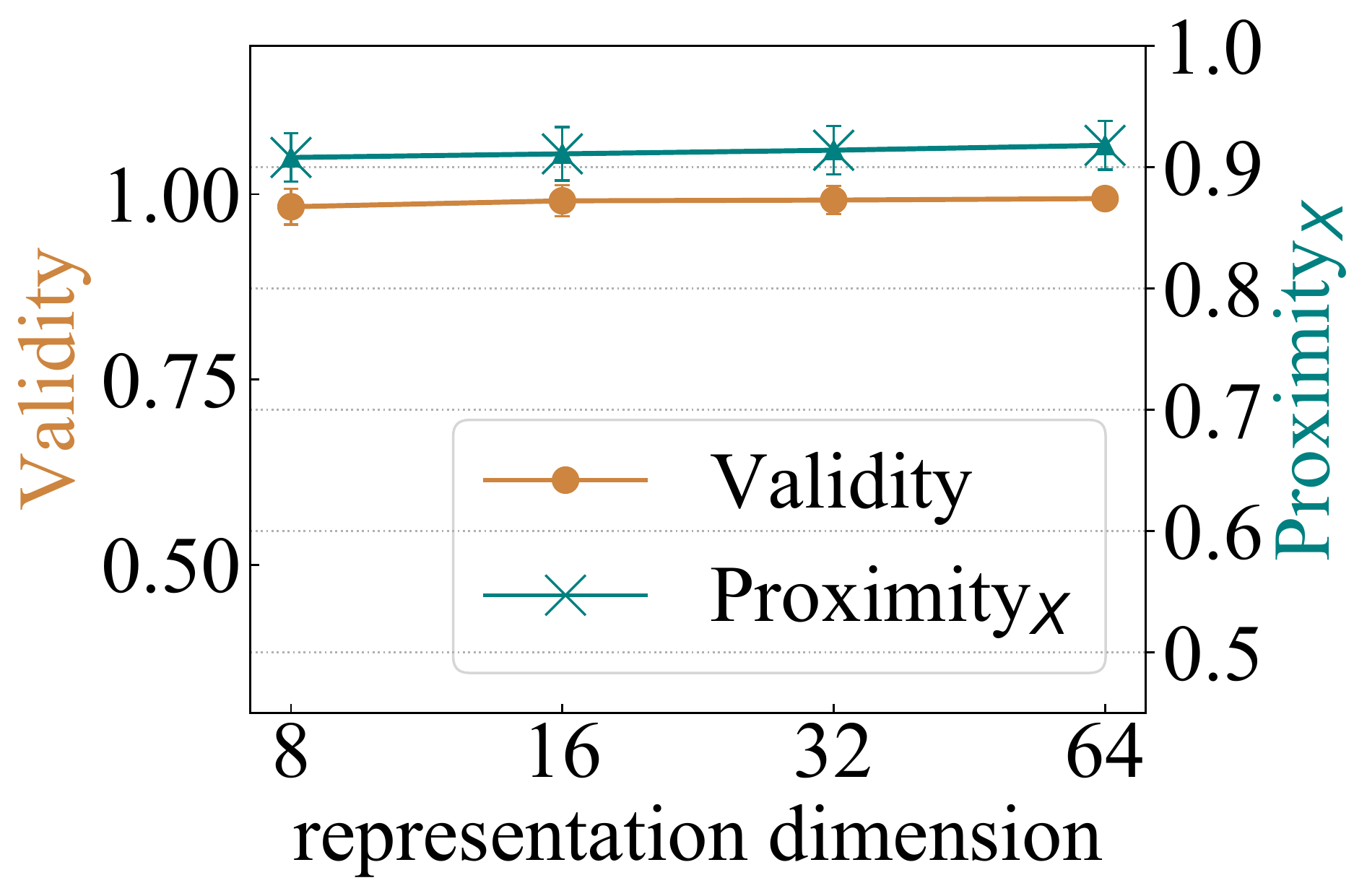}
      \vspace{-2mm}
        \caption{Representation dimension}
    \end{subfigure}
        \vspace{-2mm}
  \caption{Parameter studies on Ogbg-molhiv regarding batch size and representation dimension.}
  \label{fig:parameter_app}
  \vspace{-3mm}
\end{figure}

\subsection{Parameter Study}
Here, we conduct further parameter study with respect to the batch size and representation dimension. Specifically, we vary the batch size from range $\{100, 500, 1000, 2000\}$, and the representation dimension from range $\{8, 16, 32, 64\}$. From the results shown in Fig. \ref{fig:parameter_app}, we observe that the performance of \mymodel~under different settings of these parameters is generally stable. This observation further validates the robustness of our framework.

% More results: the representation distribution?

\section{Further Discussion}
% limitation, extension or this method
\noindent\textbf{CFEs in Other Tasks on Graphs.} In this paper, we mainly focus on the task of graph classification, but it is worth noting that the proposed framework \mymodel~can also be used for counterfactual explanations in other tasks such as node classification. More specifically, in a node classification task, \mymodel~can generate CFEs for nodes with the same loss function in Eq. (\ref{eq: loss_final}). But differently, the encoder here learns node representations instead of graph representations at the bottleneck layer. Besides, in this case, $Y^*$ is a vector which contains the desired labels for all the training nodes on graph $G$, and $S$ is the vector of auxiliary variables for all the training nodes. Notice that in a graph, nodes are often not independent. To obtain a valid counterfactual for an explainee node, not only can we change the explainee node's own features and adjacent edges, but  we can also change other nodes' features or any other part of the graph structure. Therefore, the decoder still needs to generate a graph $G^{CF}$ as a counterfactual (but this process can be more efficient, as in many scenarios, we only need to generate counterfactuals for each node's neighboring subgraph instead of the whole graph). Similarly, our framework can also be extended to generate CFEs for graphs in other tasks, such as link prediction.

%\noindent\textbf{Discussion and Future Work.} 
\noindent\textbf{Limitation, Future Work, and Negative Societal Impacts.} 
% causal model, actionablity, scalability, causal knowledge
In this work, we mainly focus on promoting optimization, generalization, and causality in counterfactual explanations on graphs, while 
other important targets (e.g., actionability  \cite{ustun2019actionable}, sparsity \cite{guidotti2018local}, diversity \cite{mothilal2020explaining}, and data manifold closeness \cite{dhurandhar2018explanations}) in traditional counterfactual explanations could be considered in graph data in the future. 
Noticeably, the definition and evaluation metrics with respect to these targets should be specifically tailored for graphs, rather than directly employed in the same way as other types of data. 
Besides, in terms of causality, another interesting direction is incorporating different levels of prior knowledge and assumptions regarding the underlying causal model into CFE generation on graphs, and quantifying the influence of  different levels of prior knowledge and assumptions on the CFE performance. 
%Especially, the graph data is often high-dimensional, and , while the causal relations between more high-level concepts are also worth consideration in CFE generation.
Currently, we have not found any negative societal impact regarding this work.
\end{document}